%% file: acl_latex.tex
\pdfoutput=1

\documentclass[11pt]{article}

\usepackage[preprint]{acl}

\usepackage{times}
\usepackage{latexsym}

\usepackage[T1]{fontenc}

\usepackage[utf8]{inputenc}

\usepackage{microtype}
\usepackage{enumitem}

\usepackage{inconsolata}

\usepackage{booktabs}
\usepackage{adjustbox}
\usepackage{enumitem}

\usepackage{amsfonts}
\usepackage{pifont}
\newcommand{\cmark}{\ding{51}}%
\newcommand{\xmark}{\ding{55}}
\usepackage{scalerel}

\usepackage{amsmath}
\usepackage{multirow}
\usepackage{algorithm}
\usepackage[
noEnd=false, 
beginComment=//~, 
endComment=, 
beginLComment=/*~, 
endLComment=~*/,
]{algpseudocodex}

\algrenewcommand\algorithmicrequire{\textbf{Input:}}
\algrenewcommand\algorithmicensure{\textbf{Output:}}

\newcommand\minipagewidth{3cm}

\title{SpaRC and SpaRP: \underline{\smash{Spa}}tial \underline{\smash{R}}easoning \underline{\smash{C}}haracterization and \underline{\smash{P}}ath Generation for Understanding Spatial Reasoning \\Capability of Large Language Models}

\author{Md Imbesat Hassan Rizvi\textsuperscript{1} \quad Xiaodan Zhu\textsuperscript{1,2} \quad Iryna Gurevych\textsuperscript{1}\\ 
  \textsuperscript{1}Ubiquitous Knowledge Processing Lab (UKP Lab), Department of Computer Science and \\
  Hessian Center for AI (hessian.AI), Technical University of Darmstadt, Germany \\
  \textsuperscript{2}Department of Electrical and Computer Engineering \& Ingenuity Labs Research Institute,  \\
  Queen’s University, Canada \\
  \textsuperscript{1}\texttt{\href{www.ukp.tu-darmstadt.de}{www.ukp.tu-darmstadt.de}} \quad \textsuperscript{2}\texttt{\href{mailto:xiaodan.zhu@queensu.ca}{xiaodan.zhu@queensu.ca}}}

\begin{document}
\maketitle
\input{Sections/abstract}
\input{Sections/introduction}
\input{Sections/related_work}
\input{Sections/methodology}
\input{Sections/experiments_and_results}
\input{Sections/conclusion}

\input{Sections/limitations_and_acknowledgements}

\bibliography{anthology,custom}

\appendix

\input{Sections/appendix}

\end{document}

%% file: Sections/abstract.tex
\begin{abstract}

Spatial reasoning is a crucial component of both biological and artificial intelligence. In this work, we present a comprehensive study of the capability of current state-of-the-art large language models (LLMs) on spatial reasoning. To support our study, we created and contribute a novel \textbf{\texttt{Spa}}tial \textbf{\texttt{R}}easoning \textbf{\texttt{C}}haracterization (\texttt{SpaRC}) framework and \textbf{\texttt{Spa}}tial \textbf{\texttt{R}}easoning \textbf{\texttt{P}}aths (\texttt{SpaRP})\footnote{\scaleobj{0.0075}{\includegraphics{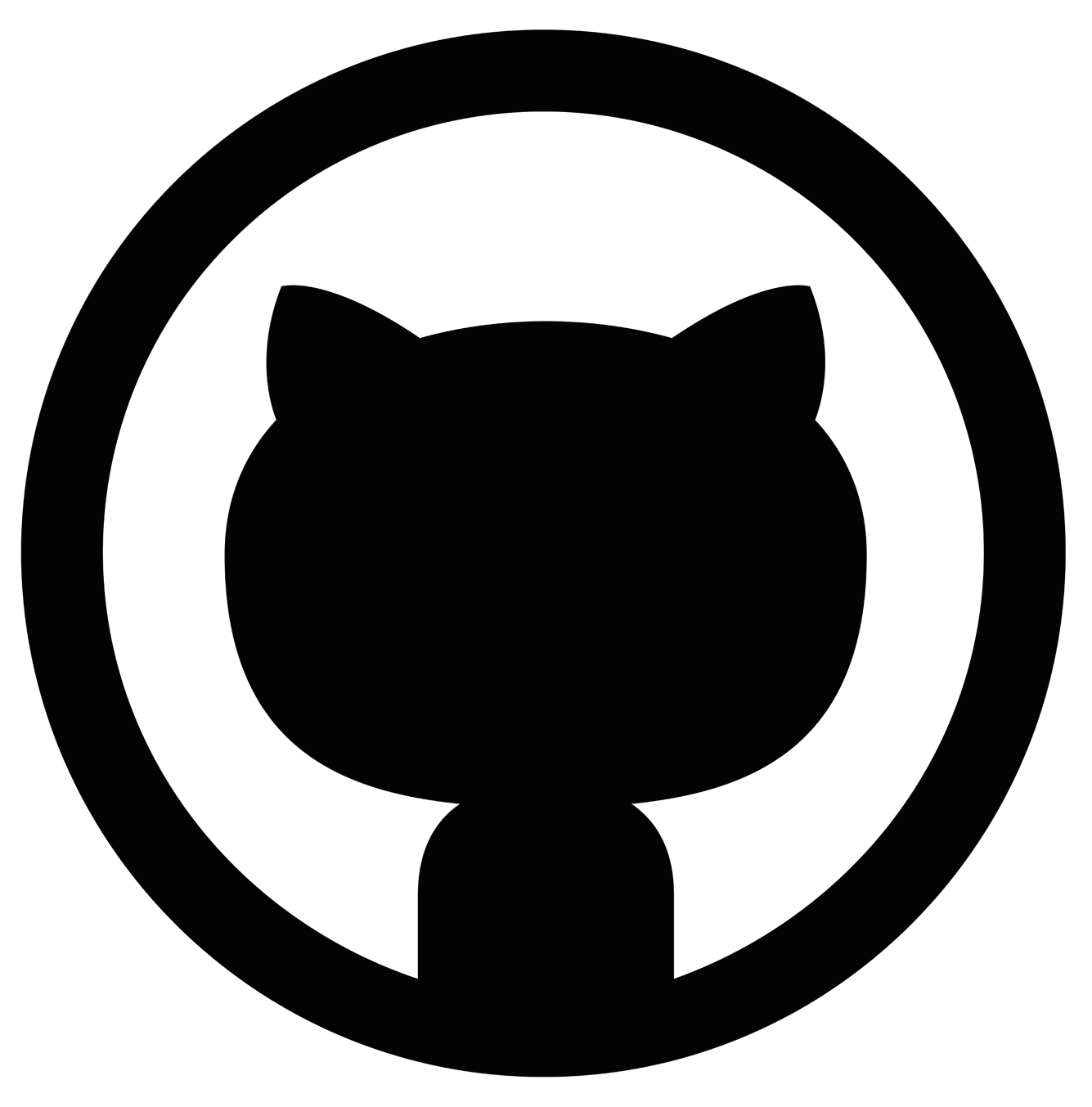}} Code: \href{https://github.com/UKPLab/acl2024-sparc-and-sparp}{https://github.com/UKPLab/acl2024-sparc-and-sparp} \\ \scaleobj{0.05}{\includegraphics{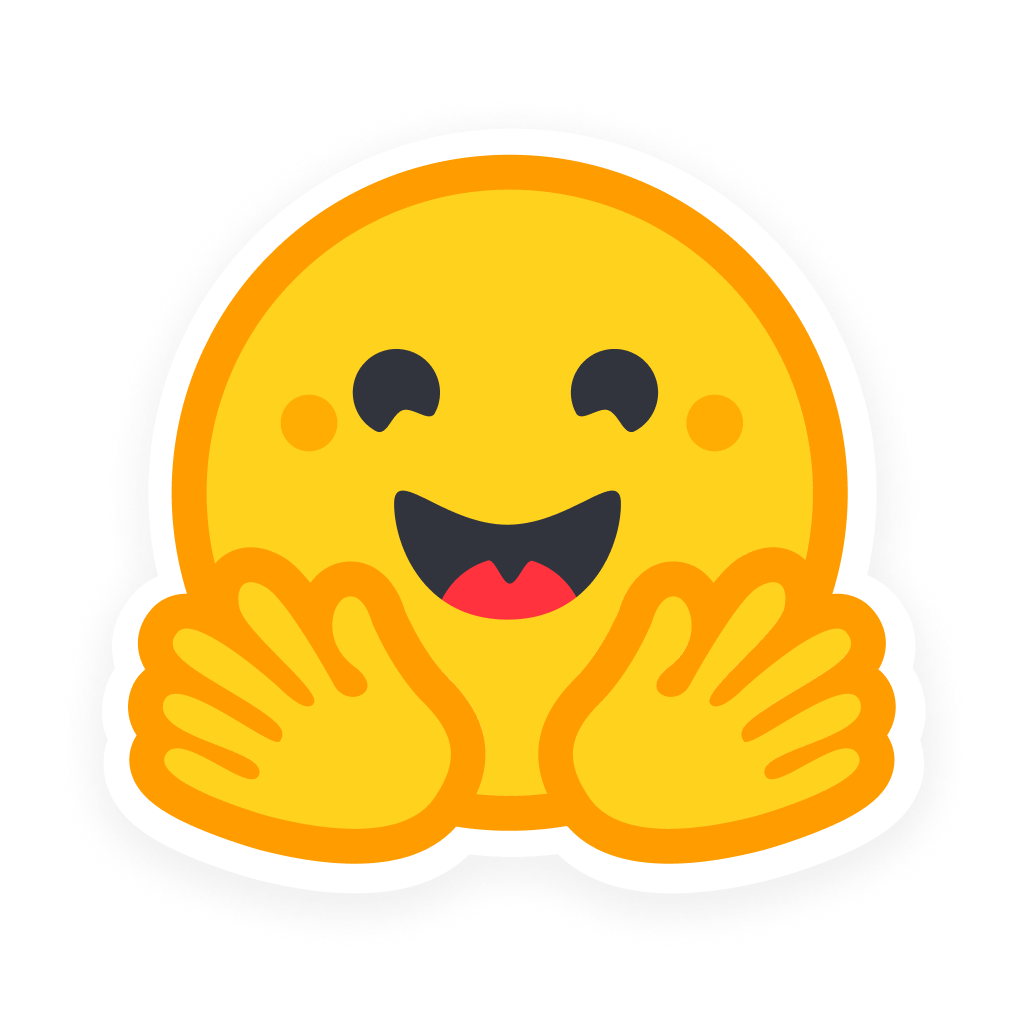}} Dataset: \href{https://huggingface.co/datasets/UKPLab/sparp}{https://huggingface.co/datasets/UKPLab/sparp} \\ TU-Datalib Dataset: \href{https://tudatalib.ulb.tu-darmstadt.de/handle/tudatalib/4235}{https://tudatalib.ulb.tu-darmstadt.de/handle/tudatalib/4235}} datasets, 
to enable an in-depth understanding of the  spatial relations and compositions as well as the usefulness of spatial reasoning chains.
We found that all the state-of-the-art LLMs do not perform well on the datasets---their performances are consistently low across different setups. The spatial reasoning capability improves substantially as model sizes scale up. Finetuning both large language models (e.g., Llama-2-70B) and smaller ones (e.g., Llama-2-13B) can significantly improve their F1-scores by 7--32 absolute points. We also found that the top proprietary LLMs still significantly outperform their open-source counterparts in topological spatial understanding and reasoning. 

\end{abstract}

%% file: Sections/introduction.tex
\section{Introduction}
\label{sec:introduction}

Spatial understanding and reasoning are a crucial component of both biological and artificial intelligence, essential for daily interactions and common tasks such as dialogues and conversations \citep{situated-dialog-2007-kruijff, language-grounding-context-2019-udagawa}, navigation \citep{r2r2018anderson,touchdown2019chen, zhang-kordjamshidi-2022-explicit}, and robotics \citep{bisk-etal-2016-natural, robot-manipulation-nli-2021-venkatesh}, among others.
They require common reasoning steps such as identifying objects, determining other objects being involved, and aggregating multiple spatial relations to reach a conclusion. The advancement of the field has significantly benefited from many well-known tasks and datasets, including 
bAbI  \citep{babi2016weston}, \textsc{SpartQA}~\citep{mirzaee-etal-2021-spartqa}, \textsc{SpaRTUN} and \textsc{ReSQ} \citep{mirzaee-kordjamshidi-2022-transfer}, and StepGame \citep{stepGame2022shi}, among others.

Recently, Large Language Models (LLMs) 
have been shown to be capable of performing abstract, commonsense-based, and multi-hop reasoning \citep{cot2022wei, llm-zero-shot-reasoners-2022kojima, selfcons2023wang}. If such models are to be used as intelligent agents to answer questions, perform tasks, and  collaborate with humans, whether they can understand the basic spatial relationships and perform corresponding reasoning would become critical to many real-life applications.

In this work, we present an extensive study on the state-of-the-art LLMs' capability in  spatial reasoning. The key components of spatial abilities include: (i) understanding spatial relations and composition, and (ii) developing reasoning chains to reach conclusions. 
Prior work \cite{mirzaee-etal-2021-spartqa, mirzaee-kordjamshidi-2022-transfer, stepGame2022shi} has focused on the relations and spatial composition tied to a limited context setup, as will be detailed later in this paper. 
In our work, we propose a bottom-up approach that builds upon detailed spatial properties, providing fine control for constructing spatial rules and context setups.
We formalize and propose  \textbf{\texttt{Spa}}tial \textbf{\texttt{R}}easoning \textbf{\texttt{C}}haracterization (\texttt{SpaRC}), a systematic framework in defining \textit{spatial properties} of objects, relations, and contexts, as well as how they \textit{characterize} spatial composition, which is inspired by the widely used benchmarks \textsc{SpaRTUN} \cite{mirzaee-kordjamshidi-2022-transfer} and StepGame \cite{stepGame2022shi}.

Reasoning paths are an integral part of the reasoning process and critical for analyzing and enhancing reasoning models. To the best of our knowledge, unlike other reasoning tasks such as mathematical reasoning, there exists no dataset with textual spatial reasoning paths. 
In this paper we develop deductively verified spatial reasoning paths by using spatial reasoners to generate step-by-step reasoning on \textsc{SpaRTUN} and StepGame, which is then verbalized to form textual chain-of-thoughts.
We show that finetuning different sizes of LLMs (13B and 70B) on the reasoning paths significantly improves their spatial reasoning performance, which also highlights the poor performance of the generalist pretrained LLMs (without finetuning) on spatial reasoning. In summary, our contributions are as follows:

\begin{itemize}[leftmargin=8pt,itemsep=-0.2em,before=\vspace{-0.1cm}]
 \item 
We present a comprehensive study on the spatial reasoning capabilities of the state-of-the-art LLMs, under extensive setups: comprehensive spatial characterizations, different parameter scales, pretrained \textit{vs.} finetuned models, and different decoding strategies. We show that the current LLMs do not perform well on the spatial reasoning tasks. We observe that spatial reasoning capability improves substantially as model sizes scale up. Top proprietary LLMs still significantly outperform their open-source counterparts in topological spatial reasoning.
    \item 
    To support an in-depth study, we present the \textbf{\texttt{Spa}}tial \textbf{\texttt{R}}easoning \textbf{\texttt{C}}haracterization (\texttt{SpaRC}) framework, a systematic bottom-up approach that shifts the focus towards spatial properties, providing a fine and flexible control on the spatial composition rules and context setups. We characterize and extend the widely used benchmark datasets \textsc{SpaRTUN} and StepGame under the \texttt{SpaRC} framework.
    \item We develop \textbf{\texttt{Spa}}tial \textbf{\texttt{R}}easoning \textbf{\texttt{P}}aths (\texttt{SpaRP}) by generating  reasoning steps using symbolic spatial reasoners and verbalizing them in a deductive step-by-step process. We demonstrate that finetuning large language models on our reasoning paths can consistently improve their spatial reasoning  abilities.
\end{itemize}

%% file: Sections/related_work.tex
\section{Related Work}
\label{sec:related_work}

\paragraph{Text-based Spatial Reasoning.} Textual spatial reasoning datasets present the task as question-answering (SRQA) over a textual spatial context. \citeauthor{babi2016weston} \citeyearpar{babi2016weston} introduced bAbI containing two datasets focused on positional (Task 17) and navigational (Task 19) reasoning. Their simplistic nature and small size prompted subsequent works to create new and challenging datasets. \citeauthor{mirzaee-etal-2021-spartqa} \citeyearpar{mirzaee-etal-2021-spartqa} designed reasoning rules, and created human-generated and synthetic context-question-answer tuples from spatial description of visual scenes (\textsc{SpartQA}) to train and evaluate spatial reasoning of neural language models. \citeauthor{mirzaee-kordjamshidi-2022-transfer} \citeyearpar{mirzaee-kordjamshidi-2022-transfer} further extended the spatial rules to cover 16 spatial relations over multiple formalisms in 3D in their synthetic \textsc{SpaRTUN} dataset, and commonsense spatial reasoning in the human-generated \textsc{ReSQ} dataset. StepGame \citep{stepGame2022shi} was introduced to assess robust positional multi-hop spatial reasoning in 2D. Our \texttt{SpaRC} framework builds on top of \textsc{SpaRTUN} and StepGame as they provide a broad coverage over the number of hops and relations for abstract spatial reasoning. 

\paragraph{Reasoning Abilities of Large Language Models.} Certain reasoning capabilities have been shown to be emergent abilities of LLMs~\citep{wei2022emergent}, which are further elicited by various chain-of-thought prompting techniques \citep{cot2022wei, llm-zero-shot-reasoners-2022kojima, tot2023liu, hao-etal-2023-reasoning}. On logic-based tasks, including spatial reasoning, they however lag behind significantly when compared to neuro-symbolic methods \citep{mirzaee-kordjamshidi-2023-disentangling, yang-etal-2023-coupling}. 

To understand spatial reasoning abilities, \citeauthor{bang-etal-2023-multitask} \citeyearpar{bang-etal-2023-multitask} provided a preliminary probing analysis on ChatGPT using a very small dataset (60 examples from each of StepGame and \textsc{SpartQA}). 
\citet{yang-etal-2023-coupling} 
evaluated the performance of GPT-3 on StepGame; \citet{mirzaee-kordjamshidi-2023-disentangling} reported the performance of GPT-3 on \textsc{SpartQA}, \textsc{SpaRTUN}, and \textsc{ReSQ} datasets. 
However, these works are limited in terms of evaluation metric, qualitative analysis, past generation of LLMs, pretrained LLMs, or generation strategies. To the best of our knowledge, our work is the first attempt at a comprehensive evaluation of spatial reasoning of LLMs under these settings.

%% file: Sections/methodology.tex
\section{The \underline{\smash{\texttt{Spa}}}tial \underline{\smash{\texttt{R}}}easoning \underline{\smash{\texttt{C}}}haracterization (\texttt{SpaRC}) Framework}
\label{sec:methodology}

The steps to identify and compose spatial relations between entities distinguish spatial reasoning from other reasoning tasks. 
Prior work e.g. \textsc{SpaRTUN} \cite{mirzaee-kordjamshidi-2022-transfer} and StepGame \cite{stepGame2022shi}, have focused directly on the spatial composition rules coupled with the contexts, which can lead to different conclusions even for the same set of relations. For example, for the same context ``A is left of B and B is above C'', applying the spatial composition of StepGame concludes that A is to the left and above C, while no directional relation between A and C can be concluded at all by applying the spatial rules of \textsc{SpaRTUN}. The conclusions are completely different \textit{but} equally valid. This difference can be reconciled by examining the underlying spatial properties of the objects and relations, specifically the treatment of objects as points vs extended, and completeness of the knowledge of relations in the context. We, therefore, advocate for an extendable bottom-up approach starting from a more granular level and introduce the \textbf{\texttt{Spa}}tial \textbf{\texttt{R}}easoning \textbf{\texttt{C}}haracterization (\texttt{SpaRC}) framework. \texttt{SpaRC} prioritizes spatial properties over spatial composition rules. Consequently, it offers finer control in creating contexts and facilitates a deeper and systematic examination of the spatial reasoning capabilities.

To keep our work closer and comparable to the widely used existing benchmarks, \textsc{SpaRTUN}~\citep{mirzaee-kordjamshidi-2022-transfer} and StepGame~\citep{stepGame2022shi}, we identify \textit{six} properties that cover and characterize these datasets by two \textit{distinct and mutually exclusive} sets of \textit{three} properties each. With \texttt{SpaRC}, we further explore two \textit{properties sets} (\texttt{PS}) with properties in common to these existing benchmarks.


\begin{table}[!ht]
\renewcommand{\arraystretch}{0.75}
    \centering
    \begin{adjustbox}{max width=\columnwidth}
    \begin{tabular}{c c c c}
        \toprule
        \textbf{
        $\mathcal{F}$} & \textbf{Sub-Type} & \textbf{Relations ($\mathcal{R}$)} & \textbf{Textual Label ($\mathcal{L}$)} \\
        \midrule
        \multirow{8}{*}{Topological} & \multirow{8}{*}{$\mathcal{T}_{R}$ (RCC8)} & \texttt{DC} & outside \\
         & & \texttt{EC} & outside and touching \\
         & & \texttt{PO} & partially overlapping \\
         & & \texttt{EQ} & overlapping \\
         & & \texttt{TPP} & inside and touching \\
         & & \texttt{NTPP} & inside \\
         & & \texttt{TPPI} & contains and touches \\
         & & \texttt{NTPPI} & contains \\
        \midrule
        \multirow{14}{*}{Directional} & \multirow{6}{*}{$\mathcal{D}_R$ (Relative)} & \texttt{LEFT} & left \\
        & & \texttt{RIGHT} & right \\
        & & \texttt{ABOVE} & above \\
        & & \texttt{BELOW} & below \\
        & & \texttt{FRONT} & front \\
        & & \texttt{BEHIND} & behind \\
        \cmidrule{2-4}
        & \multirow{4}{*}{$\mathcal{D}_C$ (Cardinal)} & \texttt{NORTH} & above \\
        & & \texttt{SOUTH} & below \\
        & & \texttt{EAST} & right \\
        & & \texttt{WEST} & left \\
        \cmidrule{2-4}
        & \multirow{4}{*}{$\mathcal{D}_T$ (Clock)} & \texttt{12 o'clock} & above \\
        & & \texttt{3 o'clock} & right \\
        & & \texttt{6 o'clock} & below \\
        & & \texttt{9 o'clock} & left \\
        \midrule
        \multirow{3}{*}{Distance} & \multirow{2}{*}{$\mathcal{S}_Q$(Qualitative)} & \texttt{NEAR} & near \\
        & & \texttt{FAR} & far \\
        \cmidrule{2-4}
        & $\mathcal{S}_U \text{(Quantitative)}$ & -- & -- \\
        \bottomrule
    \end{tabular}
    \end{adjustbox}
    \caption{Formalisms ($\mathcal{F}$) and their sub-types, relations ($\mathcal{R}$) in the datasets and their labels ($\mathcal{L}$). Labels are presented in natural language to work with language models. Composite relations e.g. lower-left are considered in a multi-label setting in the present work.}
    \label{tab:formalisms}
\end{table}

\subsection{Principle and Design of \texttt{SpaRC}}
\label{subsec:terminologies}

We focus on a set of binary spatial relations $\mathcal{R}$ (Table \ref{tab:formalisms}) by following the previous work~\citep{mirzaee-kordjamshidi-2022-transfer,stepGame2022shi}. The relations cover three formalism ($\mathcal{F}$)---topological $\mathcal{T}$, directional $\mathcal{D}$, and distance $\mathcal{S}$, divided into sub-types---region connection calculus (RCC8) $\mathcal{T}_{R}$, relative directions $\mathcal{D}_R$, cardinal directions $\mathcal{D}_C$, clock-face directions $\mathcal{D}_T$, qualitative distance $\mathcal{S}_Q$, and quantitative distance $\mathcal{S}_U$. 

For the relations set $\mathcal{R}$ and a given set of entities $\mathcal{E}$, we denote a context $\mathcal{C} = \{(h,r,t)_i\}_{i=1}^{N}$ as a set of $(h,r,t)$ tuples, where $h \in \mathcal{E}$ is a head entity, $t \in \mathcal{E}$ is the tail entity, and $r \in \mathcal{R}$ is the binary relation. Without loss of generality, objects are considered to be in a 2D space with $(x_s, y_s)$ and $(x_e, y_e)$ as the start and end positions.  
We now identify and describe six spatial properties of the objects, contexts, and relations that are crucial in determining their spatial composition rules. Refer to Appendix~\ref{sec:appendix_design_sparc} for a more detailed discussion.

\paragraph{Fixed Orientation or Point of View (FPoV).} The directional relations are considered to be axis-aligned from a fixed orientation or point of view, i.e., fixed axes in a 2D or 3D space. A fixed mapping across the relative, cardinal, and clock-face directions is usually chosen. Consistent with the prior work, we map and canonicalize cardinal $\mathcal{D}_C$ and clock-face $\mathcal{D}_T$ relations to four relative directions $\mathcal{D}_R$ (Table~\ref{tab:formalisms}), \textit{only} for their label representations $\mathcal{L}$. We denote the 2D subset of directions as $^{2D}\mathcal{D} = \mathcal{D} \setminus \{\texttt{FRONT, BEHIND}\}$.

\paragraph{Point Objects (PO).} A point object satisfies $x_s = x_e \ \land \ y_s = y_e$. As they are dimensionless, point objects have a reduced set of relations with reference to other point objects. Real objects can be treated as point objects in practical contexts when their sizes are negligible.

\paragraph{Extended Objects (EO).} 
An object is said to be an extended object if $x_s \neq x_e \ \lor \ y_s \neq y_e$. In \texttt{SpaRC}, we extend StepGame by considering extended objects in addition to point objects. We further study additional composition rules for extended objects than those presented in \textsc{SpaRTUN}, as will be detailed later in Section~\ref{subsec:data_creation}. 

\begin{table}[]
\renewcommand{\arraystretch}{0.85}
    \centering
    \begin{adjustbox}{max width=\columnwidth}
    \begin{tabular}{c c c}
        \toprule
        \textbf{Relation} & \textbf{Point Objects (PO)} & \textbf{Extended Objects (EO)} \\
        \midrule
        \underline{\textit{Incomplete (RI):}} & & \\
        \texttt{RIGHT}(A,B) & $x_A > x_B$ & $x_s^A \geq x_e^B$ \\
        \texttt{BELOW}(A,B) & $y_A < y_B$ & $y_e^A \leq y_s^B$ \\
        \midrule
        \underline{\textit{Complete (RC):}} & & \\
        \multirow{2}{*}{\texttt{RIGHT}(A,B)} & $x_A > x_B \ \land$ & $x_s^A \geq x_e^B \ \land$ \\
        & $y_A = y_B$ & $y_s^B \leq y_e^A \ \land \ y_e^B \geq y_s^A$ \\
        \cmidrule{2-3}
        \multirow{2}{*}{\texttt{BELOW}(A,B)} & $y_A < y_B \ \land$ & $y_e^A \leq y_s^B \ \land$ \\
        & $x_A = x_B$ & $x_s^A \leq x_e^B \ \land \ x_e^A \geq x_s^B$ \\
        \midrule
        \texttt{RIGHT}(A,B) $\land$ & $x_A > x_B \ \land$ & $x_s^A \geq x_e^B \ \land$ \\
        \texttt{BELOW}(A,B) &  $y_A < y_B$ & $y_e^A \leq y_s^B$ \\
        \bottomrule
    \end{tabular}
    \end{adjustbox}
    \caption{Mathematical descriptions of Relation Incomplete (RI) and Relation Complete (RC) contexts for the relations \texttt{RIGHT}, \texttt{BELOW}, and their combination in terms of entity positions ($x,y$) for Point Objects (PO) or entity boundaries ($x_s,x_e,y_s,y_e$) for Extended Objects (EO).}
    \label{tab:formal_direction_completness}
\end{table}

\begin{figure}
    \centering
    \begin{adjustbox}{max width=\columnwidth}
    \includegraphics{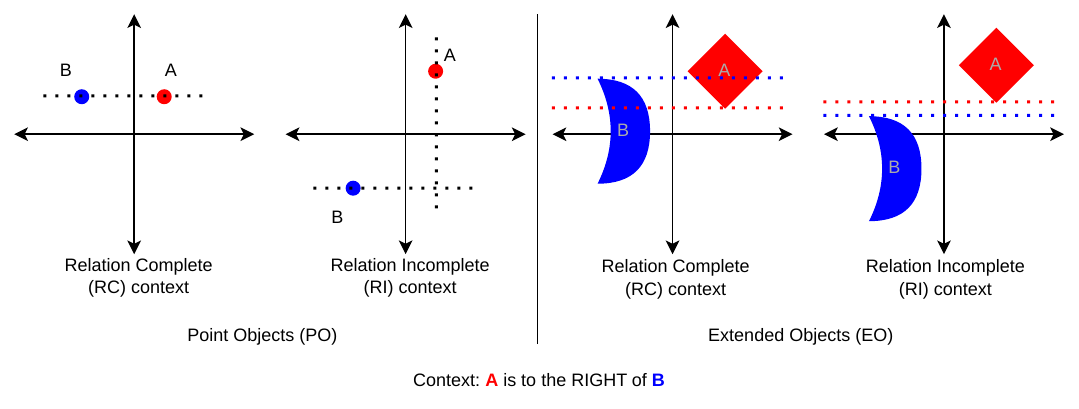}
    \end{adjustbox}
    \caption{Visualization of Relation Complete (RC) and Relation Incomplete (RI) contexts for the \texttt{RIGHT} relation for Point Objects (PO) and Extended Objects (EO).}
    \label{fig:direction_completeness}
\end{figure}

\paragraph{Relation Incomplete (RI).} We \textit{introduce} the term relation incomplete (RI) for a context $\mathcal{C}$
between a head $h$ and a tail $t$ entity if \textit{not all} the relations $r \in \mathcal{R}$ between these entities are \textit{considered} to be known and expressed in the context. 
Thus, the knowledge for the expressed relations should be \textit{treated} as incomplete or partial for spatial composition.  For example, ``Ron is to the right of Hermione'' as an RI context means that the direction orthogonal to the \texttt{RIGHT} could be \texttt{ABOVE} or \texttt{BELOW} as well. The state of positions or boundaries of objects on the \textit{orthogonal} axes cannot be assumed. 
Table~\ref{tab:formal_direction_completness} and Figure~\ref{fig:direction_completeness} exemplify and visualize this for a few scenarios.

\paragraph{Relation Complete (RC).} We \textit{introduce} the term relation complete (RC) for a context $\mathcal{C}$ 
between $h$ and $t$ if \textit{all} the relations $r \in \mathcal{R}$ between these entities are \textit{considered} to be known and expressed in the context, and treated as such for spatial compositions.
For the previous example ``Ron is to the right of Hermione'' to be considered as RC, the context should mean that Ron is only to the \texttt{RIGHT} of Hermione, and not to her lower-right or upper-right side. The positions or boundaries of objects on the \textit{orthogonal} direction axes should coincide or overlap. 
Table~\ref{tab:formal_direction_completness} and Figure~\ref{fig:direction_completeness} exemplify and visualize this for a few scenarios. In \texttt{SpaRC}, we further consider this property in conjunction with other properties, such as extended objects, to design composition rules that are not present in StepGame, as discussed later in Section~\ref{subsec:data_creation}.

We note that the presence of atomic relations, e.g., \texttt{LEFT} or composite relations, e.g., upper-left, i.e., \{\texttt{ABOVE}, \texttt{LEFT}\} in a context sentence does not necessarily imply the context to be Relation Incomplete or Relation Complete respectively. Composite relations such as upper-left can still be incomplete in 3D space or when considered along with topological relations in 2D space.

\paragraph{Quantitatively Specified (QS).} A relation which is stated in terms of a unit of measurement is said to be quantitatively specified in the given context. 
Quantitatively specified relations that are \textit{inverse} of each other, e.g. \{\texttt{LEFT}, \texttt{RIGHT}\}, can readily be composed. Consistent with StepGame, our current work considers only directional relations to be quantitatively specified in terms of distance.

\paragraph{Quantitatively Unspecified (QU).} A relation which can be stated in terms of a unit of measurement but is not stated as such in a given context is said to be quantitatively unspecified. 
Quantitatively unspecified relations that are \textit{inverse} of each other, e.g. \{\texttt{LEFT}, \texttt{RIGHT}\}, cannot be composed unless they are quantified. In \texttt{SpaRC}, we design and study the reasoning abilities for this property in conjunction with other properties, such as point objects, that are not present in \textsc{SpaRTUN} and StepGame, as discussed later in Section~\ref{subsec:data_creation}. 

\begin{table*}[]
\renewcommand{\arraystretch}{0.85}
    \centering
    \begin{adjustbox}{max width=\textwidth}
    \begin{tabular}{| l l l l l |}
        \hline
         Not & $\forall (X,Y) \in Entities $ & $R \in \{Dir \lor PP\}$ & IF $R(X,Y)$ & $\implies \text{NOT}(R_{reverse}(X,Y))$ \\ 
         Inverse & $\forall (X,Y) \in Entities$ & $R \in \{Dir \lor PP\}$ & IF $R(Y,X)$ & $\implies R_{reverse}(X, Y )$ \\
         Symmetry & $\forall (X,Y) \in Entities$ & $R \in \{Dis \lor (RCC - PP)\}$ & IF $R(Y,X)$ & $\implies R(X,Y)$ \\
         Transitivity & $\forall (X,Y,Z) \in Entities$ & $R \in \{Dir \lor PP\}$ & IF $R(X,Z), R(Z,Y)$ & $\implies R(X,Y)$ \\
         Combination & $\forall (X,Y,Z,H) \in Entities$ & $R \in Dir, \ast PP \in PP$ & IF $\ast PP(X,Z), R(Z,H), \ast PPi(H,Y)$ & $\implies R(X,Y)$ \\
         \hline
    \end{tabular}
    \end{adjustbox}
    \caption{Spatial Rules reproduced from \textsc{SpaRTUN} \citep{mirzaee-kordjamshidi-2022-transfer}. $Dir$: Directional relations (e.g., \texttt{LEFT}), $Dis$: Distance relations (e.g., \texttt{FAR}), $PP$: all Proper parts relations (\texttt{NTPP, NTPPI, TPPI, TPP}), $RCC - PP$: All RCC8 relation except proper parts relations. $\ast PP$: one of \texttt{TPP} or \texttt{NTPP}. $\ast PPi$: one of \texttt{NTPPi} or \texttt{TPPi}.}
    \label{tab:spartun_rules}
\end{table*}

We restrict our study to the above 6 properties to keep it closer and comparable to the existing benchmarks, \textsc{SpaRTUN} and StepGame. These properties form 3 \textit{mutually exclusive} pairs---\{\texttt{EO,PO}\}, \{\texttt{RI,RC}\}, \{\texttt{QS,QU}\}, leading to 8 possible sets. 
\texttt{SpaRC} 
can be extended with additional properties, however, we note that the number of possible characterizations increases exponentially with the number of properties.

\subsection{Creation of the \texttt{SpaRC} Dataset}
\label{subsec:data_creation}

We \textit{identify} the property set \texttt{PS} for the existing benchmarks, as formalized in the previous section, based on the generation process of the context and the spatial composition rules. More concretely, we identify that \textsc{SpaRTUN} is characterized by the property set 
\texttt{PS1} = \{\texttt{EO,RI,QU}\}, while StepGame is characterized by the property set 
\texttt{PS2} = \{\texttt{PO,RC,QS}\}. 
These property sets are mutually exclusive with \texttt{PS2} supporting \textit{stronger} composition rules than \texttt{PS1} for a given context, e.g. ``A is left of B and B is above C'' as discussed earlier. Refer to Appendix~\ref{sec:appendix_benchmark_characterization} for more details. 

In the \texttt{SpaRC} framework, we construct two additional datasets by relaxing the properties of StepGame from \texttt{PO} to \texttt{EO}, and \texttt{QS} to \texttt{QU}. We chose to extend StepGame as it is simple with fewer relations (only directional which is common across datasets and benchmarks) and challenging (more number of hops).  Concretely, we create the datasets \texttt{SpaRC-PS3} with the property set \texttt{PS3} = \{\texttt{PO,RC,QU}\}, and \texttt{SpaRC-PS4} with the property set \texttt{PS4} = \{\texttt{EO,RC,QU}\}. Their composition rules, elaborated upon in Section~\ref{subsec:reasoning}, are formalized by the Algorithm~\ref{alg:direction-composition} and Algorithm~\ref{alg:inequality-composition} respectively.

We confine our study to these four property sets because they encompass the two existing benchmarks, while still allowing to study the impact of additional characterizations shared with these benchmarks. We leave the extensions to further spatial characterizations and property sets as future work.

\begin{table}[!t]
\renewcommand{\arraystretch}{0.85}
    \centering
    \begin{adjustbox}{max width=\columnwidth}
    \begin{tabular}{c c c c c r r}
        \toprule
        \textbf{Dataset} & \textbf{
        $\mathcal{F}$} & \textbf{Properties} & \textbf{Textual} & \textbf{Split} & \textbf{\# Context} & \textbf{\# Ques.} \\
        & & & \textbf{Reason.} & & & \\
        \midrule
        \multirow{3}{*}{\textsc{SpaRTUN}} & \multirow{6}{*}{$\mathcal{T}_{R}$,$\mathcal{D}$,$\mathcal{S}_{Q}$} & \multirow{6}{*}{\texttt{EO,RI,QU}} & \multirow{3}{*}{\textbf{\xmark}} & Train & 6039 & 18400 \\
         &  &  &  & Dev & 915 & 2818 \\
         &  &  &  & Test & 925 & 2830 \\
        \cmidrule{4-7}
         \multirow{3}{*}{\texttt{SpaRP-PS1}} &  &  & \multirow{3}{*}{\cmark} & Train & 5806 & 16348 \\
         &  &  &  & Dev & 877 & 2392 \\
         &  &  &  & Test & 872 & 2301 \\
         \midrule
        \multirow{3}{*}{StepGame} & \multirow{6}{*}{$^{2D}\mathcal{D}$, $\mathcal{S}_{U}$} & \multirow{6}{*}{\texttt{PO,RC,QS}} & \multirow{3}{*}{\xmark} & Train & 50000 & 50000 \\
         &  &  &  & Dev & 5000 & 5000 \\
         &  &  &  & Test & 100000 & 100000 \\
         \cmidrule{4-7}
         \multirow{3}{*}{\texttt{SpaRP-PS2}} & & & \multirow{3}{*}{\cmark} & Train & 49243 & 49243 \\
         &  &  &  & Dev & 4927 & 4927 \\
         &  &  &  & Test & 98614 & 98614 \\
         \cmidrule{2-7}
         \multirow{3}{*}{\texttt{SpaRP-PS3}} & \multirow{6}{*}{$^{2D}\mathcal{D}$} & \multirow{3}{*}{\texttt{PO,RC,QU}} & \multirow{3}{*}{\cmark} & Train & 44666 & 44666 \\
         &  &  &  & Dev & 4494 & 4494 \\
         &  &  &  & Test & 78092 & 78092 \\
         \cmidrule{3-7}
         \multirow{3}{*}{\texttt{SpaRP-PS4}} &  & \multirow{3}{*}{\texttt{EO,RC,QU}} & \multirow{3}{*}{\cmark} & Train & 41436 & 41436 \\
         &  &  &  & Dev & 4171 & 4171 \\
         &  &  &  & Test & 69474 & 69474 \\
        \bottomrule
    \end{tabular}
    \end{adjustbox}
    \caption{Comparison between the extended (\texttt{SpaRP}) dataset and the source datasets. Descriptions of the properties are provided in Section~\ref{subsec:terminologies}. Relations contained in the formalisms are presented in Table~\ref{tab:formalisms}. All the questions are of Find Relations (FR) types.}
    \label{tab:dataset_comparison}
\end{table}

\begin{figure}
    \centering
    \begin{adjustbox}{max width=\columnwidth}
    \includegraphics{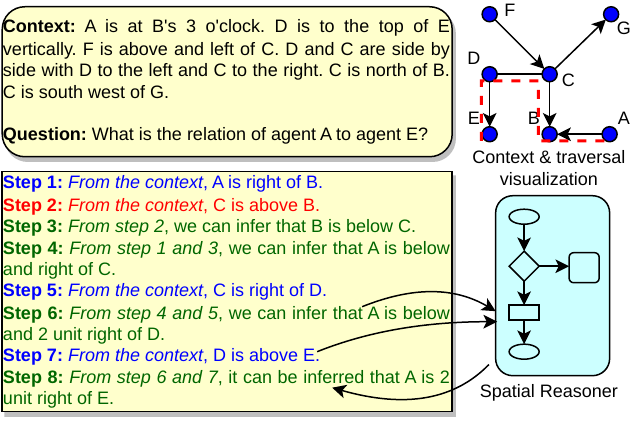}
    \end{adjustbox}
    \caption{Our step-by-step deductive \textbf{\texttt{Spa}}tial \textbf{\texttt{R}}easoning \textbf{\texttt{P}}aths (\texttt{SpaRP}) generation. A context graph and node traversal from the head to the tail entity in a question is identified and verbalized. \textcolor{blue}{Blue} indicates context relations $r^c$, \textcolor{red}{red} indicates inverse context relations $r^{ic}$, and \textcolor{ForestGreen}{green} indicates deduced relations $r^d$ between entities while traversing the reasoning path A--B--C--D--E.}
    \label{fig:reasoning_gen}
\end{figure}

\section{The \underline{\smash{\texttt{Spa}}}tial \underline{\smash{\texttt{R}}}easoning \underline{\smash{\texttt{P}}}aths (\texttt{SpaRP})}
\label{subsec:reasoning}

Reasoning paths are an integral part of reasoning models and critical for analyzing and enhancing such models. To the best of our knowledge, unlike other reasoning tasks such as mathematical reasoning, there exists no dataset with spatial reasoning paths. In this section, we develop deductively verified spatial reasoning paths by \textit{verbalizing} the symbolic steps. 

Existing spatial reasoning datasets can be considered as a collection of context-question-answer ($\mathcal{C}$, $\mathcal{Q}$, $\mathcal{A}$) tuples. Formally, we denote a context $\mathcal{C} = \{(h,r,t)_i\}_{i=1}^{N}$ defined over a set of entities $\mathcal{E}$ and binary relations $\mathcal{R}$ as a set of $(h,r,t)$ tuples, where $h \in \mathcal{E}$ is the head entity, $t \in \mathcal{E}$ is the tail entity and $r \in \mathcal{R}$ is the binary relation. For a given ($\mathcal{C, Q, A}$) tuple, seeking relation between the head $h_q$ and tail $t_q$ entities, we define a \textit{symbolic} reasoning path $\mathcal{P} = \left( l_i \right)_{i=1}^{L}$ as a sequence of $L$ reasoning links $l_i = (h_i, r^{\cup}_i, t_i)$ such that $h_1 = h_q$, $t_L = t_q$, and $h_i = t_{i-1}$ for $1 < i \leq L$. We define $r^{\cup} = r^c \cup r^{ic} \cup r^d$, where $r^c$ denotes the set of relations present in the \underline{\textbf{c}}ontext, $r^{ic}$ denotes the \underline{\textbf{i}}nverse relations present in the \underline{\textbf{c}}ontext i.e. relations from $t$ to $h$, and $r^d$ denotes the set of \underline{\textbf{d}}educed relations. Following the format of deductively verified chain-of-thought \citep{deductiveverificationcot2023ling}, we \textit{verbalize} the reasoning path $\mathcal{P}$ as a series of step-by-step reasoning sentences, where each step receives their necessary context and premises (Figure~\ref{fig:reasoning_gen}). The overall process is as given below:

\begin{algorithm}
    \caption{Relative Direction composition for set of properties \texttt{PS2} and \texttt{PS3} in 2D.}
    \label{alg:direction-composition}
    \small
    \begin{algorithmic}[1]
        \Require{Pairs to compose $\{pair1, pair2\}$.\\$quantitative \in \{true, false\}$.}
        \Ensure{$merged$ pair.}

        \LComment{initialized pair starts with $dx = dy = 0$}
        \State $merged \gets InitializePair$
        \State $merged.head \gets pair1.head$
        \State $merged.tail \gets pair2.tail$
        
        \For{$pair \in \{pair1, pair2\}$}
            \For{$delta \in \{dx, dy\}$}
                \State $delta \gets merged.delta + pair.delta$    
                \LComment{Handle direction reversal and quantitatively unspecified}
                \If{$(merged.delta \times pair.delta < 0)$ \textbf{and not} $quantitative$}
                    \LComment{Set as \textit{NaN} to invalidate compositions from now on in this direction}
                    \State $merged.delta \gets \mathit{NaN}$
                \Else
                    \State $merged.delta \gets delta$
                \EndIf
            \EndFor
        \EndFor
    \end{algorithmic}
    \normalsize
\end{algorithm}

\begin{enumerate}[leftmargin=*,itemsep=-1.8mm,before=\vspace{-2mm},after=\vspace{-2mm}]
    \item Entities and their relations in the contexts are either pre-annotated (\textsc{SpaRTUN}) or extracted using regex pattern matching (StepGame) to construct the symbolic context $\mathcal{C}$.
    \item A traversal path $\mathcal{P}$ is identified from $h_q$ to $t_q$ by constructing a network graph over $\mathcal{C}$.
    The deduced relations $r^{d}$ are initialized to be the inverse of $r^{ic}$, to traverse and merge steps in a single direction from $h_q$ to $t_q$ (Figure~\ref{fig:reasoning_gen}).
    \item We traverse the path $\mathcal{P}$, progressively merging the links (as $h_i = t_{i-1}$) and updating the deduced relations $r^d$ based on the property set \texttt{PS} and their spatial composition rules: 
    
    \begin{itemize}[leftmargin=8pt,itemsep=0em,before=\vspace{-2mm}]
        \item For 
        \textsc{SpaRTUN} we reuse the rules from 
        \citet{mirzaee-kordjamshidi-2022-transfer}, 
        reproduced in Table~\ref{tab:spartun_rules}.
        \item For 
        StepGame and 
        \texttt{SpaRC-PS3}, 
        we represent the relative 
        positions as signed integers on the $x$ and $y$ axis, and numerically compose them (Algorithm~\ref{alg:direction-composition}). Without the quantitative knowledge of backtracking along a given axis, e.g. $x$-axis for \{\texttt{LEFT}, \texttt{RIGHT}\}, no subsequent inferences can be made for those directions.
        \item For 
        \texttt{SpaRC-PS4}, 
        the relations in context can be expressed as logical conjunction $\land$ of inequalities, refer to Section~\ref{sec:methodology}, Table~\ref{tab:formal_direction_completness}, and Figure~\ref{fig:direction_completeness}. For composition of relations to merge reasoning steps, consistency of inequalities for relations $r \in \mathcal{D}$ is checked and the deduced relations set $r^d$ is updated (Algorithm~\ref{alg:inequality-composition}).
    \end{itemize}
    \item We finally \textit{verbalize} the reasoning path $\mathcal{P}$ link-by-link (Figure~\ref{fig:reasoning_gen}) following the format of deductively verified chain-of-thought \citep{deductiveverificationcot2023ling}. However, instead of generating and self-verifying LLM outputs, we use spatial reasoners
    for ground truth generation. 
\end{enumerate}

\begin{algorithm}[!t]
    \caption{Relative Direction composition for set of properties \texttt{PS4} in 2D.}
    \label{alg:inequality-composition}
    \small
    \begin{algorithmic}[1]
        \Require{Pairs to compose $\{pair1, pair2\}$.\\current set of constraint inequalities $ineq$}
        \Ensure{$merged$ pair and updated inequalities $ineq$.}

        \LComment{initialize an empty pair}
        \State $merged \gets InitializePair$
        \State $merged.head \gets pair1.head$
        \State $merged.tail \gets pair2.tail$

        \For{$rel \in \{\texttt{LEFT, RIGHT, ABOVE, BELOW}\}$}
            \State $candidate\_ineq \gets \texttt{substitute\_entities}($
            \State \quad \quad $rel.ineq, \ merged.head, \ merged.tail)$
            \State $consistent \gets \texttt{check\_consistency}($
            \State \quad \quad $candidate\_ineq, \ ineq)$
            \If{$consistent$}
                \State $\texttt{insert}(candidate\_ineq, \ ineq)$
                \State $\texttt{insert}(rel, \ merged.relations)$
            \EndIf
        \EndFor
        
    \end{algorithmic}
    \normalsize
\end{algorithm}

We denote the extended dataset as \textbf{\texttt{Spa}}tial \textbf{\texttt{R}}easoning \textbf{\texttt{P}}aths (\texttt{SpaRP}). Specifically, we extended \textsc{SpaRTUN}, StepGame, \texttt{SpaRC-PS3}, and \texttt{SpaRC-PS4}, to be \texttt{SpaRP-PS1}, \texttt{SpaRP-PS2}, \texttt{SpaRP-PS3} and \texttt{SpaRP-PS4}, respectively, by enriching the former with the reasoning paths.
A comparison of the derived datasets with the original datasets is summarized in Table~\ref{tab:dataset_comparison}. 

%% file: Sections/experiments_and_results.tex
\section{Experimental Setup}
\label{sec:experiments}

\paragraph{Dataset.} Due to the expense and resource limitations for running LLMs, for each of the four subsets of \texttt{SpaRP}, we randomly sample 2000, 500, and 1000 datapoints as our training, validation, and test set, respectively. We call them small \texttt{SpaRP}, or \textbf{\texttt{SpaRP-S}}. We also randomly sample equal number of instances for each \textit{number of hops} in the reasoning path. Additionally, we collect \textit{five} diverse sets of \textit{human-generated} natural language descriptions of the properties relevant to spatial compositions, and construct a \textit{system prompt} template with a unified task instruction using these descriptions. 

\paragraph{Implementation Details.} To help replicability, we include implementation details such as dataset sampling, system prompt, and training parameters in Appendix-\ref{sec:appendix_experiment}.

\paragraph{Evaluation Metrics.} We use exact-match accuracy and macro-averaged F1-scores\footnote{We used the \href{https://pypi.org/project/scikit-learn/}{scikit-learn} v1.3.2 library.}.

\begin{table}[!t]
\renewcommand{\arraystretch}{0.9}
    \centering
    \small
    \begin{adjustbox}{max width=\columnwidth}
    \begin{tabular}{c c l c c}
        \toprule
        \multicolumn{2}{c}{\textbf{Dataset}} & \textbf{Model} & \textbf{Acc.} & \textbf{F1} \\
        \midrule
         \multirow{8}{*}{
         {\rotatebox[origin=c]{90}{\texttt{SpaRP-S-PS1}}}}
         & \multirow{8}{*}{{\rotatebox[origin=c]{90}{(\textsc{SpaRTUN})}}}
         & Llama-2-13B & 0.2 & 0.49 \\
         & & Llama-2-13B-FT & 18.9 & 22.23 \\
         \cmidrule(l{0.5em}r{0.5em}){3-5}
         & & Llama-2-70B & 10.1 & 23.37 \\
     & & Llama-2-70B-FT & 28 & 36.49 \\
         & & Llama-2-70B$_{\text{SC=20}}$ & 17.1 & 27.95 \\
         \cmidrule(l{0.5em}r{0.5em}){3-5}
         & & GPT-4 & 46.8 & 54.30 \\
         & & GPT-4$_{\text{SC=20}}$ & \textbf{54.3} & \textbf{60.32} \\
         \cmidrule(l{0.5em}r{0.5em}){3-5}
         & & SOTA (\textsc{PistaQ}) & \textit{94.52} & -- \\
        \midrule
          \multirow{8}{*}{
         {\rotatebox[origin=c]{90}{\texttt{SpaRP-S-PS2}
         }}}
         & \multirow{8}{*}{{\rotatebox[origin=c]{90}{(StepGame)}}} & Llama-2-13B & 0.1 & 0.47 \\
         & & Llama-2-13B-FT & 13.7 & 33.23 \\
         \cmidrule(l{0.5em}r{0.5em}){3-5}
         & & Llama-2-70B & 10.6 & 26.41 \\ 
        & & Llama-2-70B-FT & 16.6 & 34.63 \\
        & & Llama-2-70B$_{\text{SC=20}}$ & 20.30 & 38.96 \\
         \cmidrule(l{0.5em}r{0.5em}){3-5}
        & & GPT-4 & 23.9 & 41.09 \\
        & & GPT-4$_{\text{SC=20}}$ & \textbf{28.6} & \textbf{43.01} \\
         \cmidrule(l{0.5em}r{0.5em}){3-5}
        & & SOTA (\textsc{LLM-ASP}) & \textit{90.88} & -- \\
        \midrule
        \multirow{7}{*}{
         {\rotatebox[origin=c]{90}{\texttt{SpaRP-S-PS3}}}}
        & & Llama-2-13B & 0.2 & 0.92 \\
        &  & Llama-2-13B-FT & 27.3 & 32.01 \\
        \cmidrule(l{0.5em}r{0.5em}){3-5}
        & & Llama-2-70B & 9.4 & 25.27 \\
        & & Llama-2-70B-FT & 19.5 & 32.97 \\
        & & Llama-2-70B$_{\text{SC=20}}$ & 15.2 & 32.01 \\
         \cmidrule(l{0.5em}r{0.5em}){3-5}
        & & GPT-4 & 23.8 & 35.17 \\
        & & GPT-4$_{\text{SC=20}}$ & \textbf{32.5} & \textbf{42.06} \\
        \midrule
        \multirow{7}{*}{
         {\rotatebox[origin=c]{90}{\texttt{SpaRP-S-PS4}}}}
        & & Llama-2-13B & 0.7 & 1.84 \\
        & & Llama-2-13B-FT & 30.6 & 31.62 \\
        \cmidrule(l{0.5em}r{0.5em}){3-5}
        & & Llama-2-70B & 9.0 & 22.13 \\
        & & Llama-2-70B-FT & 20 & 31.74 \\
        & & Llama-2-70B$_{\text{SC=20}}$ & 18.3 & 29.73 \\
         \cmidrule(l{0.5em}r{0.5em}){3-5}
        & & GPT-4 & 21.7 & 33.02 \\
        & & GPT-4$_{\text{SC=20}}$ & \textbf{32.9} & \textbf{40.23} \\
        \bottomrule
    \end{tabular}
    \end{adjustbox}
    \caption{Performance evaluations of Llama-2 (13B and 70B) and GPT-4 models on the spatial reasoning datasets. SC=20 means self-consistency over 20 generations, and FT indicates finetuned model with greedy decoding.}
    \label{tab:evaluation}
\end{table}

\section{Results and Analysis}
\label{sec:results}

We run experiments with three state-of-the-art LLMs  --- Llama-2-13B, Llama-2-70B \cite{touvron2023llama}, and GPT-4\footnote{ The default GPT-4, specifically GPT-4-0613, used in the experiments was accessed between December 1, 2023, and January 31, 2024.}, each one with both single greedy decoding and self-consistency \citep{selfcons2023wang} with majority voting over 20 generations with sampling (SC=20). Inputs are provided with a ``system prompt'' containing task instructions and \textit{5-shot} CoT with randomly sampled exemplars from the \textit{relevant dev-set}, e.g., exemplars for a test instance of \texttt{SpaRP-S-PS1} (\textsc{SpaRTUN}) were randomly sampled from its own dev-set. We also finetune Llama-2 13B and 70B models, indicated by FT in Table~\ref{tab:evaluation}, using QLoRA~\cite{qlora2023dettmers} on the \textit{verbalized reasoning paths} made available by \texttt{SpaRP}.

\paragraph{Overall Results.} 
As shown in Table~\ref{tab:evaluation}, we observe that the performance of all the state-of-the-art LLMs on the spatial reasoning datasets is low, lagging significantly behind the existing state-of-the-art symbolic-based models such as \textsc{PistaQ} \citep{mirzaee-kordjamshidi-2023-disentangling} and LLM-ASP \citep{yang-etal-2023-coupling} on \textsc{SpaRTUN} and StepGame, respectively. This suggests that if these generalist models are to be used for any spatial-reasoning-related tasks (e.g., in LLMs-based agents), caution should be exerted.

Among these models, GPT-4 under SC=20 exhibits the best performance overall, followed closely by GPT-4 with greedy decoding. The latter outperforms even the largest open-source Llama-2-70B model with SC=20. 

We also observed that the spatial reasoning ability of LLMs improves significantly with increasing model sizes. The smaller pre-trained Llama-2 13B model essentially exhibits no spatial reasoning ability, with the F1-scores of 0.49, 0.47, 0.92, and 1.84 on \texttt{SpaRP-S-PS1} (\textsc{SpaRTUN}), \texttt{SpaRP-S-PS2} (StepGame), \texttt{SpaRP-S-PS3}, and \texttt{SpaRP-S-PS4}, respectively. In contrast, the larger pre-trained Llama-2 70B model demonstrates \textit{comparatively significant} spatial reasoning ability, achieving F1-scores of 23.37, 26.41, 25.27, and 22.13 on \texttt{SpaRP-S-PS1} (\textsc{SpaRTUN}), \texttt{SpaRP-S-PS2} (StepGame), \texttt{SpaRP-S-PS3}, and \texttt{SpaRP-S-PS4}, respectively.

\begin{figure}
    \centering
    \begin{adjustbox}{max width=\columnwidth}
    \includegraphics{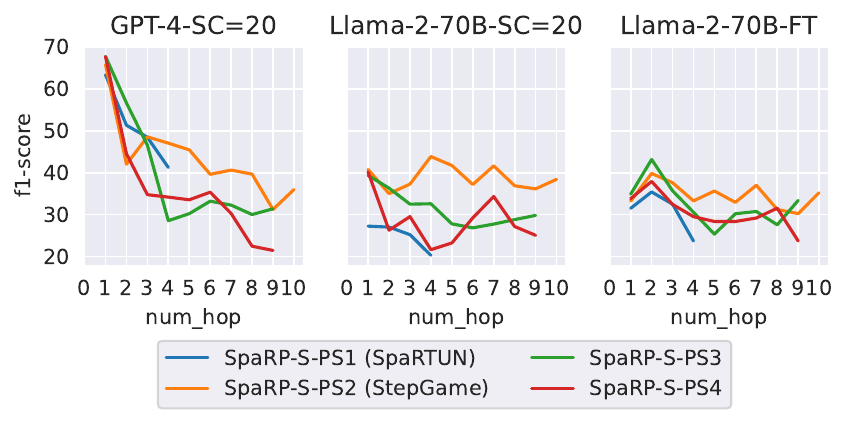}
    \end{adjustbox}
    \caption{F1 scores vs. ground truth number of hops for spatial reasoning across the datasets and models. SC=20 means self-consistency over 20 generations, and FT indicates finetuned model with greedy decoding.
    }
    \label{fig:f1scores_hops}
\end{figure}

\begin{figure}
    \centering
    \begin{adjustbox}{max width=\columnwidth}
    \includegraphics{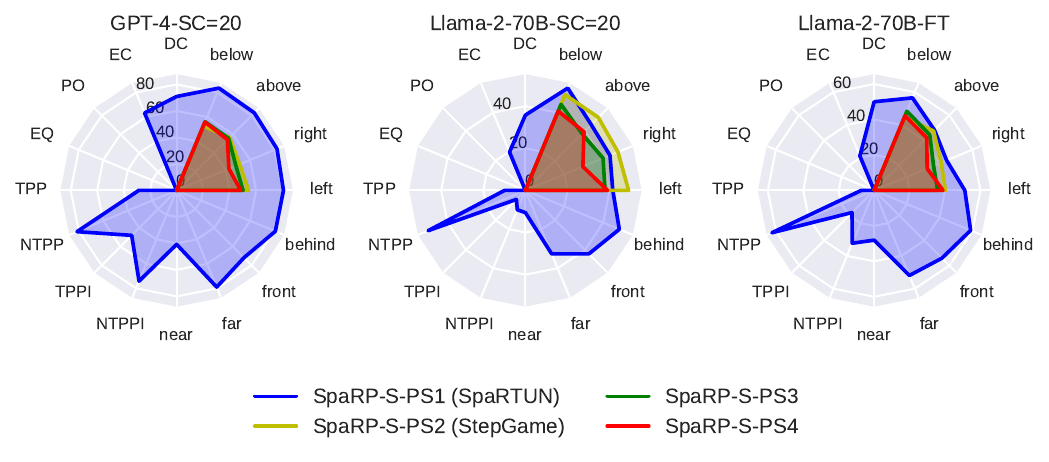}
    \end{adjustbox}
    \caption{F1 scores of individual labels across the datasets and models. SC=20 means self-consistency over 20 generations, and FT indicates finetuned model with greedy decoding.
    }
    \label{fig:f1scores_labels}
\end{figure}

\paragraph{Impact of Spatial Properties and Composition Rules.} 
StepGame and \texttt{SpaRP-S-PS3} consider entities as point objects (PO), however, \texttt{SpaRP-S-PS3} does not quantify directions rendering them incomposable while backtracking, e.g. \texttt{RIGHT} followed by \texttt{LEFT} is not composable. \texttt{SpaRP-S-PS4} considers entities as real objects with extended sizes, thereby introducing added complexity to spatial relation composition (Section~\ref{subsec:reasoning} and Algorithm~\ref{alg:inequality-composition}). The F1-scores (Table~\ref{tab:evaluation}) of both GPT-4 and Llama-2 underscore these challenges. 

Furthermore, Figure~\ref{fig:f1scores_hops} demonstrates that the F1-scores of both \texttt{SpaRP-S-PS3} and \texttt{SpaRP-S-PS4} consistently trail those of \texttt{SpaRP-S-PS2} (StepGame) across varying numbers of hops. This highlights the utility of our \texttt{SpaRC} framework in identifying additional challenges that are not addressed by the existing benchmarks.

\paragraph{Relation-wise Performance.} The performance of GPT-4 is significantly better compared to Llama-2 models on \texttt{SpaRP-S-PS1} (\textsc{SpaRTUN}), which has a larger candidate set comprising of 16 relations, including 8 topological relations. In contrast, \texttt{SpaRP-S-PS2} (StepGame) has a smaller candidate set consisting of only directional relations. This highlights a notable deficiency in Llama-2 regarding the understanding and composition of topological relations. More importantly, even the finetuned Llama-2 model falls short of GPT-4's performance. The top proprietary LLMs still significantly outperform their open-source counterparts in topological spatial reasoning.

Additionally, Figure~\ref{fig:f1scores_hops} demonstrates that even when controlling for the same number of hops, the F1-scores of Llama-2 on \texttt{SpaRP-S-PS1} (\textsc{SpaRTUN}) rank lowest across all hops. An examination of F1-scores on a per-relation basis (Figure~\ref{fig:f1scores_labels}) further confirms this difficulty of topological relations for Llama-2 models compared to GPT-4. 

\begin{table}[!t]
    \centering
    \begin{adjustbox}{max width=\columnwidth}
    \begin{tabular}{l c c}
    \toprule
    \multirow{2}{*}{\textbf{Dataset}} & \multicolumn{2}{c}{\textbf{Pearson correlation coefficients ($\rho$)}} \\
    \cmidrule{2-3}
    & Llama-2-70B-FT & GPT-4 \\
    \midrule
    \texttt{SpaRP-S-PS1} (\textsc{SpaRTUN}) & 0.535 & 0.775 \\
    \texttt{SpaRP-S-PS2} (StepGame) & 0.299 & 0.414 \\
    \texttt{SpaRP-S-PS3} & 0.375 & 0.409 \\
    \texttt{SpaRP-S-PS4} & 0.446 & 0.352 \\
    \bottomrule
    \end{tabular}
    \end{adjustbox}
    \caption{Pearson correlation coefficients ($\rho$) between the observed number of hops in the model generated output and the ground truth number of hops.}
    \label{tab:hops_corrcoef}
\end{table}

\renewcommand{\minipagewidth}{\columnwidth}
\begin{table}[]
\renewcommand{\arraystretch}{0.8}
    \centering
    \small
    \begin{adjustbox}{max width=\columnwidth}
    \begin{tabular}{l}
        \toprule
        \multicolumn{1}{c}{\textbf{Errors, examples and explanations}} \\
        \midrule
        \begin{minipage}{\minipagewidth} \textbf{Error:} Incorrect relation extraction \\
        \textbf{Context:} \textbf{Box EEE} \textit{\textcolor{ForestGreen}{has}} a \textbf{tiny white rectangle} and covers a midsize white diamond. \\
        \textbf{Extracted:} \textit{Step 7:} It is given that the \textbf{tiny white rectangle} is \textit{\textcolor{red}{inside and touching}} the \textbf{box EEE}. \\
        \textbf{Explanation:} \textit{Has} only means inside.\end{minipage} \\
        \midrule
        \begin{minipage}{\minipagewidth} \textbf{Error:} Reverse answer \\
        \textbf{Question:} What is the relation of the agent \textbf{W} to the agent \textbf{X}? \\
        \textbf{Answer Step:} \textit{Step 8:} From step 5 and 7, we can infer that \textbf{X} is above and left of \textbf{W}. \textit{Hence, the answer is} above, and left. \\
        \textbf{Explanation:} Directional relations are non-symmetric. Question is from W to X, while answer is from X to W.\end{minipage} \\
        \midrule
        \begin{minipage}{\minipagewidth} \textbf{Error:} Copied, not composed \\
        \textbf{Reasoning Steps:} \textit{Step 6:} From step 4 and 5, we can infer that \textbf{A} is below and right of \textbf{S}. \textit{Step 7:} From the context, \textbf{S} is \textcolor{red}{left} of \textbf{M}. \textit{Step 8:} From step 6 and 7, we can infer that \textbf{A} is below and right of \textbf{M}. \\
        \textbf{Explanation:} Relation from S to M not used in composition. Instead, relation from A to S is copied in step 8.\end{minipage} \\
        \midrule
        \begin{minipage}{\minipagewidth} \textbf{Error:} Composed without connection \\
        \textbf{Reasoning Steps:} \textit{Step 5:} From step 3 and 4, we can infer that \textbf{Y} is right of \textbf{L}. \textit{Step 14:} From step 12 and 13, we can infer that \textbf{C} is below and right of \textbf{K}. \textit{Step 15:} From step 5 and 14, we can infer that \textbf{C} is below and right of \textbf{L}. \\
        \textbf{Explanation:} No common entity between merged steps 5 and 14 which are 9 steps apart.\end{minipage} \\
        \bottomrule
    \end{tabular}
    \end{adjustbox}
    \caption{Errors, their examples (only relevant steps) and explanations in the model generated reasoning paths.}
    \label{tab:errors}
\end{table}

\paragraph{Finetuning with Reasoning Paths.}  
We observe that finetuning the 13B and 70B models with the reasoning paths made available in \texttt{SpaRP} consistently improves the spatial reasoning capabilities. Finetuning consistently boosts the F1-score by 21--32 and 7--13 points for 13B and 70B models respectively, across the datasets. The finetuned models exhibit significantly improved performance compared to self-consistency for \texttt{SpaRP-S-PS1} (Table~\ref{tab:evaluation}). Figure~\ref{fig:f1scores_labels} illustrates that this is primarily due to the improvements in the identification and reasoning of the topological and qualitative distance-based relations. Topological relations, such as ``inside \textit{vs} inside and touching'' or ``contains \textit{vs} contains and touches'', that differ only in terms of connectedness are often difficult for models to differentiate during identification due to the connectedness (``touch'') being either implicitly specified in the context or implicitly assumed by the models. In contrast, the performance of finetuned \textit{vs} self-consistency based generation is comparable across \texttt{SpaRP-S-PS2} to \texttt{SpaRP-S-PS4}, which are direction-only datasets. However, inference with finetuned models would still be preferable as they are computationally less intensive. Moreover, finetuning is required for smaller models with limited reasoning capabilities (e.g., 13B), where self-consistency may not be feasible.

Finally, the accuracy of a finetuned 13B model, in specific instances, surpasses that of 5-10 times larger models such as Llama-2-70B with SC=20, and GPT-4. We hope the proposed reasoning-path generation can be further used for improving LLMs' explainability and robustness on spatial reasoning.

\paragraph{Error Analysis of Reasoning Paths.} We observe that GPT-4 follows the expected decrease in performance with increasing number of hops, more consistently, compared to Llama-2 models (Figure~\ref{fig:f1scores_hops}). We attribute this to the difference between the \textit{ground truth} num\_hop ($x$-axis in Figure~\ref{fig:f1scores_hops}) and the num\_hop \textit{observed} in the model generated output. This relationship is underscored by the Pearson correlation coefficient ($\rho$) between the observed and the ground truth num\_hop as presented in Table~\ref{tab:hops_corrcoef}. The correlation coefficient of Llama-2 model, notably for \texttt{SpaRP-S-PS2} (StepGame), lags significantly when compared to GPT-4, resulting in a more erratic trend (Figure~\ref{fig:f1scores_hops}).

We sampled and manually analyzed a total of 80 model generated reasoning paths across all datasets for both the GPT-4 and Llama-2 70B models. The deductive step-by-step reasoning path made available by \texttt{SpaRP} proves to be useful in identifying errors in the generated outputs (Table~\ref{tab:errors}). Commonly observed errors include incorrect parsing or retrieval of relations from the contexts, especially for topological relations. Additionally, we observe instances of reverse answering, where relations between tail to head entities are returned instead of head to tail entities in a question. More complex reasoning failures involve copying relations from one of the reasoning steps instead of composing them. Similarly, composing relations between reasoning steps without a common entity is observed frequently over distant steps. 
Additional errors with examples are provided in Appendix~\ref{sec:appendix_errors}. These errors are more prevalent in Llama-2 models, resulting in poorer performance compared to GPT-4.

%% file: Sections/conclusion.tex
\section{Conclusion}
\label{sec:conclusion}

Spatial reasoning is one of the basic components of intelligence. We perform a  study on the spatial reasoning abilities of the latest LLMs under comprehensive setups. To support the study, we introduce (\texttt{SpaRC}), a systematic framework to characterize spatial reasoning scenarios by identifying and defining six spatial properties of objects, spatial relations, and contexts, and their impact on the spatial composition rules. Based on that, we create the (\texttt{SpaRP}) reasoning paths for the datasets.
We found that the state-of-the-art LLMs do not perform well on the datasets --- their performances are consistently low across different setups. The spatial reasoning capability improves significantly as model sizes scale up. Finetuning both large language models (e.g., Llama-2-70B) and smaller ones (e.g., Llama-2-13B) can significantly improve their performance by 7--32 points on F1-scores.  
We also found top proprietary LLMs still significantly outperform their open-source counterparts in topological spatial understanding and reasoning. We provide detailed analyses and insights in our experiments.

%% file: Sections/limitations_and_acknowledgements.tex
\section*{Limitations}

We aimed to characterize various properties of the objects, relations, contexts and the associated spatial composition rules. We, however, note that the spatial scenarios, relations and interactions between objects can still be incomplete. Further, the existing datasets and our extensions of them still pertain to a limited combination of the characterizations \textit{in isolation} in a context. Even with our proposed characterizations, a combination of these within a single context is common in the real world, including multi-modality with visual perception, which we haven't considered in our current study. The base datasets, although textual, are synthetic in nature. Combined with the use of symbolic reasoners for our reasoning path generation, our dataset inherit all the associated limitations such as relative lack of linguistic diversity, types of objects, relations etc. Finally we note that due to the cost and resource constraints of using LLMs, we worked with a smaller set of 1000 test instances per dataset, which is a common data size to work with LLMs.

\section*{Acknowledgements}

This work has been funded by the Collaboration Lab with Nexplore ``AI in Construction'' (AICO). We gratefully acknowledge the support of Microsoft with a grant for access to OpenAI GPT models via the Azure cloud (Accelerate Foundation Model Academic Research).

We also express our gratitude to Furkan Şahinuç, Chen Cecilia Liu, Vivek Khetan, and Thy Thy Tran to provide natural language descriptions of the spatial properties and characterizations that were used as part of the system prompt for the LLMs. We further thank our anonymous reviewers and Irina Bigoulaeva, Andreas Waldis, and Haishuo Fang for their fruitful discussions and helpful feedback.

%% file: Sections/appendix.tex
\section{Additional details and comparison of spatial properties in \texttt{SpaRC}}
\label{sec:appendix_design_sparc}

A symbolic context $\mathcal{C} = \{(h,r,t)_i\}_{i=1}^{N}$ is usually verbalized as several natural language sentences. However, we note that the \textit{verbalization} can be a conjunction of multiple tuples in a single context sentence e.g. ``Objects A and B are inside the box C'', or ``Entity X is below and left of entity Y''. Such verbalization is common in existing benchmarks, including \textsc{SpaRTUN} and StepGame.

\paragraph{Fixed Orientation or Point of View (FPoV).} The relations are considered to axis-aligned from a globally fixed orientation or point of view, i.e., fixed axes in a 2D or 3D space. We note that the cardinal ($\mathcal{D}_C$) and clock-face ($\mathcal{D}_T$) directions have only 4 relations in 2D. With the set of relative directions ($\mathcal{D}_R$) being larger (6 relations in 3D), $\mathcal{D}_C$ and $\mathcal{D}_T$ are mapped and canonicalized to four of the relative directions \textit{only} for their label representations $\mathcal{L}$ (Table~\ref{tab:formalisms}). Their understanding in the contexts and questions is still required. Additionally, the understanding of the map to a canonicalized label is also required to return correct answers.

\paragraph{Point Objects (PO) vs Extended Objects (EO).} Point objects are entities that are either dimensionless i.e. their boundaries on all axes coincide, or can be treated as such in a given context. Since they are dimensionless, in relation to other point objects, the possible topological $\mathcal{T}_R$ relation (Table \ref{tab:formalisms}) collapses just to \{\texttt{DC}, \texttt{EQ}\} i.e. outside or ``disconnected'', and overlapping respectively. When combined with other formalisms such as directional relations ($\mathcal{D}$), even \texttt{DC} becomes redundant as the presence of any directional relation implies that the objects are not at the same position. Although point objects are purely mathematical constructs, real objects can often be treated as point objects in practical contexts. For example when the sizes of the objects can be ignored in relation to the distances between them. e.g. discussing spatial (directional) relations between buildings across several towns.

Extended Objects, on the other hand, are entities that are not dimensionless, i.e.  their boundaries on at least one axis extends or has a spread. All real objects are extended objects. Dimensions of objects cannot be ignored when the distances between them are comparable to their sizes for spatial rule compositions and thus they must be treated as extended objects e.g. ``a number of curious silver instruments'' standing on Dumbledore's ``spindle-legged tables''.

\paragraph{Relation Incomplete (RI) vs Relation Complete (RC).} For a set of relations $\mathcal{R}$, the contexts are usually relation incomplete in several real-world scenarios or when $|\mathcal{R}|$ is relatively large. On the other hand, the contexts can be relation complete in the real-world scenarios if $|\mathcal{R}|$ is relatively small, and one needs to emphasize and be specific.

\paragraph{Quantitatively Specified (QS) vs Quantitatively Unspecified (QU).} For our current set of formalism (Table~\ref{tab:formalisms}), some topological relations $r \in \mathcal{T} \setminus$ \{\texttt{EC, EQ, TPP, TPPI}\} and all the directional relations $r \in \mathcal{D}$ can be quantitatively specified. However, the topological relations are usually considered qualitatively, although there are metric based calculus for RCC8 and other topological relations as well. For example, context statements ``Hogwarts is 200 miles to the left of the Azkaban Fortress'' and ``The Quidditch Stadium is inside and 1 KM away from the Hogwarts School's northern boundary'' have \texttt{LEFT} and \texttt{NTPP} (inside) as quantitatively specified relations respectively. Quantitatively specified relations that are \textit{reverse} of each other, such as \texttt{LEFT} and \texttt{RIGHT}, can readily be composed. For example, we can infer that Harry is 2 unit right of Ron, from the context statements -- Harry is 3 unit left of Hermione, and Hermione is 5 unit right of Ron. Relations are quantitatively specified when their measurements are required in a context directly, or to infer other spatial relations indirectly. 

On the other hand, for the previous examples, the context statements ``Hogwarts is to the left of the Azkaban Fortress'' and ``The Quidditch Stadium is inside the Hogwarts School'' are quantitatively unspecified for the relations \texttt{LEFT} and \texttt{NTPP} (inside) respectively. Quantitatively unspecified relations that are \textit{reverse} of each other, such as \texttt{LEFT} and \texttt{RIGHT}, cannot be composed unless the relations are quantified. For example, directional relation between Harry and Ron cannot be determined from the context statements -- Harry is left of Hermione, and Hermione is right of Ron.

\paragraph{Mutual Exclusitivity of Spatial Relations.} While the \textit{reverse} relations in any formalism cannot occur simultaneously, under RCC8 calculus, multiple topological relations $\mathcal{T}_R$ cannot occur simultaneously for the same \textit{ordered} pair of entities even if they are not \textit{reverse} of each others. Thus, for a given relation $r \in \mathcal{T}_R$ and an ordered pair of entities $(X, Y)$:

$r(X,Y) \implies \text{NOT}(r'(X,Y)) \ \forall r' \in \mathcal{T}_R \setminus r$

For example, \texttt{TPP} (inside and touching) and \texttt{NTPP} (inside) are exclusive in RCC8. Stating a single topological relation in $\mathcal{T}_R$ makes the context Relation Complete (RC) in (\textit{and only in}) $\mathcal{T}_R$. 

However, negative implications are only for \textit{reverse} relations in directional formalism $\mathcal{D}$. Orthogonal relations such as \texttt{LEFT} and \texttt{ABOVE} can be \textit{simultaneously true} for a set of \textit{ordered} pair of entities. As directional relations are not symmetric, we will always mean an ordered pair or sequence of entities while discussing them, unless stated otherwise. Hence, Relation Incomplete (RI) contexts can be quite common in terms of directional relations.


\section{Characterization of \textsc{SpaRTUN} and StepGame}
\label{sec:appendix_benchmark_characterization}

\begin{figure}
    \centering
    \begin{adjustbox}{max width=\columnwidth}
    \includegraphics{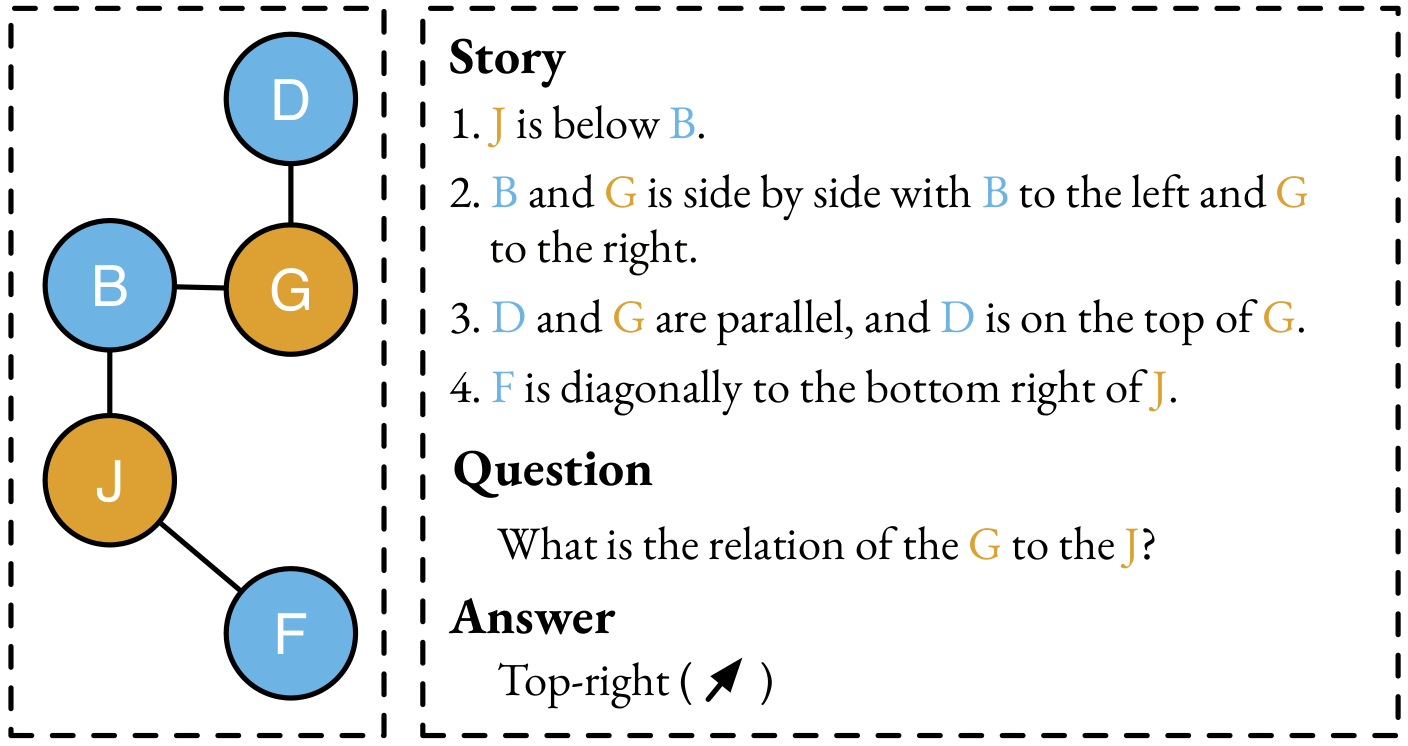}
    \end{adjustbox}
    \caption{An example reproduced from the StepGame \cite{stepGame2022shi}.}
    \label{fig:stepgame-diagram}
\end{figure}

Although the existing datasets, inlcuding \textsc{SpaRTUN} and StepGame, do not explicitly consider the spatial properties, their contexts and spatial composition rules conform to a set of these properties referenced in Section~\ref{subsec:terminologies}. StepGame considers entities in a completely abstract sense placed on a grid (Figure~\ref{fig:stepgame-diagram}). They support only directional relations (including composites such as lower-left) and an overlap. Hence, objects can either be completely overlapping or completely separate. Their placement on the grid is also in terms of implicit unit of measurements. An overlap and unrestricted numerical composition of directions during their generation process coupled with the complete abstract representation of the entities essentially make them to be point objects (PO) and quantitatively specified (QS). Additionally their clear and complete expressions such as ``BB is to the right of AA'', ``BB is at the 3 o'clock position relative to AA'', and ``AA and BB are horizontal and AA is to the right of BB'' all considered equivalent means that when the relation is expressed as \texttt{RIGHT}, it means exactly and only \texttt{RIGHT} and this relation is completely known and they correspond to the relation complete (RC) context. This is why they support \textit{strong} compositions for example presented at the beginning of Section~\ref{sec:methodology} -- ``A is left of B and B is above C'' $\implies$ ``A is to the left and above C''. Hence, the properties set of StepGame is \{\texttt{PO, RC, QS}\}.

\textsc{SpaRTUN} on the other hand considers objects that have shapes and sizes as they built their dataset on top of NLVR images and scene graphs with different sizes of objects and blocks, and support of topological relations such as containment, inside etc. Hence, their entities are extended objects (EO). Their spatial rules (Table~\ref{tab:spartun_rules}) also do not consider quantitative relations either explicitly or implicitly. Finally their spatial rules also do not make any assumption about the alignment of directional relations to be exactly parallel to an axis system. That is why a statement such as ``A is to the left of B'' doesn't rule out the possibility of A additionally being above or below B i.e. the relations are not necessarily only as stated and other directional relations would still needs to be checked rather than assumed to be not present. This is in contrast with StepGame. Thus, the properties set of \textsc{SpaRTUN} is \{\texttt{EO, RI, QU}\}. This is why applying their spatial rules (Table~\ref{tab:spartun_rules}) lead to no conclusion for the previous example ``A is left of B and B is above C'' presented at the beginning of Section~\ref{sec:methodology}. \textsc{SpaRTUN} composition rules are thus weaker in comparison to StepGame's composition based on these differences in their properties sets.


\section{Implementation Details}
\label{sec:appendix_experiment}

\subsection{Datasets and Prompts}
\label{sec:exp_dataset}

\renewcommand{\minipagewidth}{\textwidth}
\newcommand{\termwidth}{0.1\textwidth}
\begin{table*}[]
    \centering
    \small
    \begin{adjustbox}{max width=\textwidth}
    \begin{tabular}{c c}
        \toprule
        \textbf{Terminology} & \textbf{Descriptions} \\
        \midrule
        \begin{minipage}{\termwidth} System Instruction Template \end{minipage} & \begin{minipage}{\minipagewidth} You are an expert assistant with the knowledge of spatial relations and the rules to compose them under the assumptions that the contexts provided are of `\{point\_of\_view\_type\}', the objects or entities are to be treated as `\{entity\_type\}', the directions are `\{quantitative\_type\}', and `\{relation\_type\}'. The description of these terminologies are as given below:\\
        
        \{point\_of\_view\_type\}: \{point\_of\_view\_type\_desc\}\{point\_of\_view\_type\_default\} \\
        
        \{entity\_type\}: \{entity\_type\_desc\}\{entity\_type\_default\} \\
        
        \{quantitative\_type\}: \{quantitative\_type\_desc\}\{quantitative\_type\_default\} \\
        
        \{relation\_type\}: \{relation\_type\_desc\}\{relation\_type\_default\} \\
        
        You need to identify the sub-set of entities from the context that are relevant as well as combine their spatial relations with valid compositions under the above mentioned assumptions to find the spatial relations between the entities in the asked questions. The list of all possible spatial relations are: \{spatial\_relation\_choices\}. Always provide the final answer, only and only, in terms of these spatial relations. Include all the spatial relations that hold true as the answer, in case of multiple correct choices. \end{minipage}\\
        \midrule
        \begin{minipage}{\termwidth} Fixed Orientation Point of View \end{minipage} & \begin{minipage}{\minipagewidth} The spatial relations are expressed from a single, consistent and unchanging perspective. This means that the observations are made from a global viewpoint that remains same and constant for all the entities in a given context. Hence, relations such as relative directions e.g. left or right always refer to the same directions and there is a one-to-one mapping between relative, cardinal and clock-face directions i.e. left is same as west or 9 o'clock position, right is same as east or 3 o'clock position, above is same as north or 12 o'clock position, and below is same as south or 6 o'clock position. \end{minipage} \\
        \midrule
        \begin{minipage}{\termwidth} Implicit Quantification \end{minipage} & \begin{minipage}{\minipagewidth} Unless otherwise stated, consider the direction relations specified in the context to be of 1 unit distance. For example, the sentence, entity X is to the lower-left of entity Y means that the entity X is 1 unit to the left and 1 unit below the entity Y. \end{minipage} \\
        \bottomrule
    \end{tabular}
    \end{adjustbox}
    \caption{Human-generated natural language descriptions for common terminologies, defaults and system instruction. Terms inside \{\} are placeholders that are further replaced with their language descriptions. For current work, point\_of\_view\_type is always Fixed Orientation Point of View and the only default available is for quantitative\_type = Quantitatively Specified (QS) with quantitative\_type\_default = Implicit Quantification. All other placeholders are replaced by randomly sampled descriptions from one of their 5 diverse human-generated descriptions presented in Table~\ref{tab:po_descriptions} through Table~\ref{tab:qu_descriptions}. The spatial\_relation\_choices are the relevant labels $\mathcal{L}$ from Table~\ref{tab:formalisms}.}
    \label{tab:term_descriptions}
\end{table*}

\renewcommand{\minipagewidth}{\textwidth}
\begin{table*}[]
    \centering
    \small
    \begin{adjustbox}{max width=\textwidth}
    \begin{tabular}{p{\textwidth}}
        \toprule
        \multicolumn{1}{c}{\textbf{Diverse human-generated Descriptions for Point Objects (PO)}} \\ 
        \midrule
        \begin{minipage}{\minipagewidth} \textbf{Description 1:} Two objects can be treated as Point Objects in a given context for specifying their spatial relations if they are extremely small such that their sizes are immaterial, or if they are of similar or even varying shapes and sizes but are placed sufficiently far enough that their shapes and sizes can be ignored to state and compose spatial relations between them. This leads to a limitation on the spatial relations that can be specified between objects e.g. containment, but simpler relation compositions since shapes and sizes of the objects need not be considered. For example, a tea-cup and an apple on a table, or a school building and a warehouse that are miles away can be considered as point objects. \\
        \textbf{Description 2:} While composing spatial relations between objects, they can be considered as Point Objects if they can be treated as dimensionless i.e. if (1) their sizes are so small that they can be neglected or (2) the size and shape of the objects are negligible compared to the great distance between the objects. Although this situation may prevent to express certain relations like containment, it provides simpler spatial relation statements and compositions over multiple objects, since the size and shape are not considered. For example, two balls on the basketball pitch or two buildings that are separated with 2 KM distance. \\
        \textbf{Description 3:} Point Objects are small objects in a given context, whose sizes and shapes can be ignored. Thus, only their locations and orientations are considered when specifying spatial relations, leading to less number of relations and their simpler combinations over objects. A typical example of point objects can be buildings on a map or beads on a table. \\
        \textbf{Description 4:} In this scenario, objects can be treated as Point Objects if they are extremely small or far apart to the extent that their shapes and sizes can be ignored. In such cases, certain spatial relationships, like containment, become inapplicable. Additionally, since the shapes and sizes of the objects are not important, relationship compositions can be simpler. For example, two cars that are miles apart can be considered as point objects. \\
        \textbf{Description 5:} Entities can be treated as Point Objects when the distance between them relative to their sizes is either large or can be ignored. Therefore when providing spatial relations between them, a limited set of relations with simpler composition rules is possible. For example, when someone says, a cafe and a house that are far apart can be treated as point objects. \end{minipage} \\
        \bottomrule
    \end{tabular}
    \end{adjustbox}
    \caption{Five diverse human-generated natural language descriptions of Point Objects (PO).}
    \label{tab:po_descriptions}
\end{table*}

\renewcommand{\minipagewidth}{\textwidth}
\begin{table*}[]
    \centering
    \small
    \begin{adjustbox}{max width=\textwidth}
    \begin{tabular}{p{\textwidth}}
        \toprule
        \multicolumn{1}{c}{\textbf{Diverse human-generated Descriptions for Extended Objects (EO)}} \\ 
        \midrule
        \begin{minipage}{\minipagewidth} \textbf{Description 1:} Two objects are to be treated as Extended Objects in a given context for specifying their spatial relations if their shapes and sizes in comparison to the distances between them can not be ignored to state and compose spatial relations between them. This leads to more number of possible spatial relations that can be specified between objects e.g. containment, but reduces the number while increasing the complexity of possible relation compositions, as the shapes and sizes of the objects can neither be assumed nor be discarded. For example, a tea-cup and a tube-light, or a table and a cupboard in a room are to be considered as extended objects. \\
        \textbf{Description 2:} If the distance between objects is comparable to the shapes and sizes of the objects while specifying the spatial relations, the objects are considered as Extended Objects i.e. they can't be treated as dimensionless and they have significant length, breadth or height in comparison to the distances between the objects in the context. Although this gives an opportunity to use more specific spatial relations like touching or containment, the complexity of compositions increases. A basket and an apple in it or two entities, X and Y, in a room can be given as examples. \\
        \textbf{Description 3:} Extended Objects refer to objects, whose shapes and sizes can affect the spatial relations that can be specified and the way they can be combined between objects. This leads to more number of relations and the combination of relations have to be minimal in the absence of the information about the shape and size of the objects. Examples of extended objects include buildings on a street or boxes in a room. \\
        \textbf{Description 4:} In this scenario, two objects are considered to be Extended Objects if their shapes and sizes, in comparison to the distances between them, cannot be ignored. In such cases, a larger set of spatial relations between objects can be specified, although the relation composition becomes more limited when the shapes and sizes of the objects are unknown compared to when this information is known. For example, a tea-cup and a lamp or a sofa and a TV in a room can be considered as extended objects. \\
        \textbf{Description 5:} Entities can be treated as Extended Objects if they have shapes and sizes which are not to be ignored in the context. Because of this, although a larger set of relations is possible between objects but the composition rules can become complex. For example, a cafe and a mall building can be treated as extended objects and the cafe can be a part of i.e. inside the mall building itself. \end{minipage} \\
        \bottomrule
    \end{tabular}
    \end{adjustbox}
    \caption{Five diverse human-generated natural language descriptions of Extended Objects (EO).}
    \label{tab:eo_descriptions}
\end{table*}

\renewcommand{\minipagewidth}{\textwidth}
\begin{table*}[]
    \centering
    \small
    \begin{adjustbox}{max width=\textwidth}
    \begin{tabular}{p{\textwidth}}
        \toprule
        \multicolumn{1}{c}{\textbf{Diverse human-generated Descriptions for Relation Incomplete (RI) contexts}} \\ 
        \midrule
        \begin{minipage}{\minipagewidth} \textbf{Description 1:} Not all set of possible spatial relations that hold true between two objects are stated while specifying the relations between those objects. Thus, there could be multiple possible spatial configurations that conform to the stated relations between the objects. For example, the statement, object A is to the left of object B, when considered as relation incomplete could mean that A may or may not be strictly only to the left of B, i.e. it can be either only to the left, or is to the left and above, or is to the left and below B. \\
        \textbf{Description 2:} Although some spatial relations between two objects exist, they might be overlooked while expressing the relations between those objects. Therefore, other valid configurations, which are compatible with the expression but not explicitly specified, may also exist. For instance, the relation incomplete expression, the entity X is to the left of the entity Y does not have to mean that X is to the left of Y and they are strictly aligned at the same time. The entity X can be both to the left and bottom (or above etc.) of the entity Y. \\
        \textbf{Description 3:} An incomplete spatial relationship corresponds to the insufficient information or context to decide the exact spatial relationship between objects, leading to ambiguation. In other words, there can be multiple valid spatial arrangements or layouts that hold true to each incomplete relation. For example, given the incomplete statement that box `one' is in front of box `two', it holds true for multiple arrangements such as box `one' is to the right and front of box `two', or box `one' is to the left and front of box `two'. \\
        \textbf{Description 4:} Relations are incomplete in the context statements if not all the spatial relationships that exist between two objects are stated. In such cases, multiple spatial outline or positioning of the objects are possible, without a single definitive truth. For example, consider the relationship - the fruit F is behind the object O in a 2D plane. Although O is in front of F, their relative position on the horizontal axis is incomplete, and hence, could be left, right or at the same place when considered horizontally. \\
        \textbf{Description 5:} The provided set of spatial relations between two objects may not be enough to communicate the complete spatial position between them. Therefore, for the provided spatial information between two objects more than one arrangement is possible. For example, a metal ball is hanging below a metal beam in the workshop - can mean various spatial positions such as the metal ball is below the beam, or additionally, it can be to the right or left and away from the beam in consideration.\end{minipage} \\
        \bottomrule
    \end{tabular}
    \end{adjustbox}
    \caption{Five diverse human-generated natural language descriptions of Relation Incomplete (RI).}
    \label{tab:ri_descriptions}
\end{table*}

\renewcommand{\minipagewidth}{\textwidth}
\begin{table*}[]
    \centering
    \small
    \begin{adjustbox}{max width=\textwidth}
    \begin{tabular}{p{\textwidth}}
        \toprule
        \multicolumn{1}{c}{\textbf{Diverse human-generated Descriptions for Relation Complete (RC) contexts}} \\ 
        \midrule
        \begin{minipage}{\minipagewidth} \textbf{Description 1:} All set of possible spatial relations that hold true between two objects are stated while specifying the relations between those objects. Hence, there is only one spatial configuration that conforms to the stated relations between the objects. For example, the statement, object A is to the left of object B, when considered as complete could only and only mean that A is to the left of B. \\
        \textbf{Description 2:} All existing spatial relations between two objects are included while expressing the relations. Therefore, there is one-to-one mapping between spatial configuations and expressed spatial relations between objects. For instance, a relation complete statement, the entity X is to the right of the entity Y means that X is to the right of Y and they are aligned. \\
        \textbf{Description 3:} Completely specified spatial relations refer to the complete sets of spatial relations that can be held as well as stated between objects. Thus, there can be only one valid spatial arrangement or layout that holds true for a relation complete statement. An example is that box `one' is in front of box `two' and they are in the same line that denotes front in a given fixed orientation for all. \\
        \textbf{Description 4:} Relations are complete in a setting, if all the spatial relationships between two objects are stated. In such cases, there is a single ground truth spatial outline or positioning of the objects. For example, consider the relationship - the fruit F is behind the object O in a 2D plane. This means that O is strictly and only in front of F and are aligned on the axis i.e. can be considered to be neither left nor right but at the same position on the horizontal axis. \\
        \textbf{Description 5:} The provided set of relations between two objects are enough to know the actual spatial position between them. Therefore, no additional information is needed to understand the actual position between two objects. For example, a metal ball is hanging below a metal beam in the workshop means that the ball is below the beam and not to its left or right. \end{minipage} \\
        \bottomrule
    \end{tabular}
    \end{adjustbox}
    \caption{Five diverse human-generated natural language descriptions of Relation Complete (RC).}
    \label{tab:rc_descriptions}
\end{table*}

\renewcommand{\minipagewidth}{\textwidth}
\begin{table*}[]
    \centering
    \small
    \begin{adjustbox}{max width=\textwidth}
    \begin{tabular}{p{\textwidth}}
        \toprule
        \multicolumn{1}{c}{\textbf{Diverse human-generated Descriptions for Quantitatively Specified (QS) relations}} \\ 
        \midrule
        \begin{minipage}{\minipagewidth} \textbf{Description 1:} Spatial relations, such as directions, specified between two objects are said to be Quantitatively Specified if those relations can have a unit of measurement and are also stated, implicitly or explicitly, in the specified context. The composition of such relations is always possible when all the object parameters and the relations between any two objects in a statement are completely known. For example, with constraints such as objects A, B and C are apples lying in a line and the relation specified are of 1 unit measurement when not mentioned explicitly, the quantitatively specified statements - B is 3 units to the left of A, and C is to the right of B - can lead to the only conclusion that A is 2 units to the right of C, or its inverse equivalent i.e. C is 2 units to the left of A. \\
        \textbf{Description 2:} Unit of measurements in spatial relations (e.g., directions) between two objects needs to be explicitly or implicitly specified for these relations to be called as Quantitatively Specified. The composition of such relations can be determined when all other object parameters and relations of two objects are given. For example, let entities X and Z be perfect round shaped balls. Let entity Y be a round basket with 10 unit radius and let centers of all objects are horizontally aligned. If X is 1 unit to the left of the center of Y and Z is 2 units to the left of X, then Z is inside the basket and 3 units to the left of the center of the basket Y. \\
        \textbf{Description 3:} Spatial relations are Quantitatively Specified when these relations are defined with a specific unit of measurement such as meters or miles. The relation compositions over objects become deterministic if all the other object parameters and the relationships between them are provided. For example, box `one' is 3 units above box `two' and they are in the same line can be easily used to determine relations with respect to a third box, say box `three', if its position is also quantitatively specified with one of them. \\
        \textbf{Description 4:} Under this setting, spatial relations between two objects are said to be Quantitatively Specified if the relations have a unit of measurement and stated directly or indirectly in the context. In such cases, when all the object parameters and relations between any two objects in the statement are known, a deterministic relation composition is possible. For example, although there are limitations like having three apples (A1, A2, A3) arranged in a row, the statements - A2 is 2 units left of A1, and A3 is 1 unit right of A2 - provides enough information to determine the exact positions of A1 and A3 relative to each other. \\
        \textbf{Description 5:} If the quantitative value along with the measurement unit for a spatial relation is provided then those relations are said to be Quantitatively Specified. The measurements may be a default value that is understood in the context or is explicitly provided. The composition of these relations will result in a distinctly resolved relation. For example, in the sentence, the cafe is 2 blocks north of my house and the hospital is 1 block south of the cafe, it can be easily determined that the hospital is 1 block north of my house. \end{minipage} \\
        \bottomrule
    \end{tabular}
    \end{adjustbox}
    \caption{Five diverse human-generated natural language descriptions of Quantitatively Specified (QS) relations.}
    \label{tab:qs_descriptions}
\end{table*}

\renewcommand{\minipagewidth}{\textwidth}
\begin{table*}[]
    \centering
    \small
    \begin{adjustbox}{max width=\textwidth}
    \begin{tabular}{p{\textwidth}}
        \toprule
        \multicolumn{1}{c}{\textbf{Diverse human-generated Descriptions for Quantitatively Unspecified (QU) relations}} \\ 
        \midrule
        \begin{minipage}{\minipagewidth} \textbf{Description 1:} Spatial relations, such as directions, specified between two objects are said to be quantitatively unspecified if those relations can have a unit of measurement but are not stated in the specified context. The composition of such relations may not be possible even when all the object parameters and the relations between any two objects in a statement are completely known. For example, even with constraints such as objects A, B and C are apples lying in a line, the quantitatively unspecified statements - B is to the left of A, and C is to the right of B - can not lead to any conclusion regarding left, right, or overlapping relationship between A and C. \\
        \textbf{Description 2:} In order for spatial relations between two objects to be considered as quantitatively unspecified, unit of measurement in these relations should not be specified. The exact composition or realization of such relations may not be determined even if the other object features and relations are completely known. For example, let entity X be in the basket Y of a known and stated size, and let entity Z be to the right of the entity X. It is not possible to infer whether entity Z is in the basket Y or not if its distance from X is quantitatively unspecified. \\
        \textbf{Description 3:} Spatial relations are Quantitatively Unspecified when these relations are not defined in terms of specific units of measurement such as meters or miles. The relation compositions over objects can still not be determined even if all the other object parameters and the relationships between them are provided. For example, if box `one' is above box `two', it's not clear how far exactly box `one' lies with respect to box `two' and this will affect the conclusions to be drawn about relations with respect to other objects, say box `three'. \\
        \textbf{Description 4:} In this setting, spatial relations between two objects are Quantitatively Unspecified if the relations have a unit of measurement that is not specified in the context. In such cases, even when all the object parameters and relations between any two objects in the statement are known, a deterministic composition of relations may be impossible. In this scenario, although there are limitations like having three apples (A1, A2, A3) arranged in a row, the statements that lack specific quantities - A2 is on the left of A1, and A3 is on the right of A2 - do not provide enough information to determine the left, right, or overlapping positions of A1 and A3 relative to each other. \\
        \textbf{Description 5:} If the quantitative value along with the measurement unit for a spatial relation is not provided then those relations are said to be quantitatively unspecified. The composition of these relations may not be enough to result in a distinctly resolved relation. For example, in the sentence, the cafe is to the north of my house and the hospital is to the south of the cafe, it can't be determined if the hospital is to the south or north of my house. \end{minipage} \\
        \bottomrule
    \end{tabular}
    \end{adjustbox}
    \caption{Five diverse human-generated natural language descriptions of Quantitatively Unspecified (QU) relations.}
    \label{tab:qu_descriptions}
\end{table*}

We created the \textbf{\texttt{SpaRP-S}} dataset with train, validation, and test splits of sizes 2000, 500, and 1000 respectively for each sub-dataset of \texttt{SpaRP}. To ensure a \textit{fair} distribution of the difficulty level, we \textit{randomly} sample \textit{equal} number of instances for each \textit{number of hops} (of entities) in the reasoning path. The final distribution is still skewed due to less number of instances for higher number of hops in \textsc{SpaRTUN}. Additionally, we collect \textit{five} diverse sets of \textit{human-generated} natural language descriptions of the properties (Table~\ref{tab:term_descriptions}) relevant to spatial compositions (Section~\ref{subsec:terminologies}). We construct a \textit{system prompt} template with a unified task instruction and populate it with randomly sampled natural language descriptions for each instances of each sub-dataset. The system prompt template and the human-generated descriptions are presented in Table~\ref{tab:term_descriptions} through Table~\ref{tab:qu_descriptions}.

\subsection{Model configurations and training setup}
\label{sec:exp_models}

To assess the spatial understanding and reasoning abilities of LLMs over varying model sizes, we run experiments with three model variants (all chat versions) -- Llama-2-13B, Llama-2-70B, and GPT-4. The default GPT-4, specifically GPT-4-0613, used in the experiments was accessed between December 1, 2023, and January 31, 2024. 

We finetune a single model 13B and 70B models on all the four datasets i.e. \texttt{SpaRP-S-1} (\textsc{SpaRTUN}), \texttt{SpaRP-S-PS2} (StepGame), \texttt{SpaRP-S-PS3}, and \texttt{SpaRP-S-PS4}. For finetuning, we used QLoRA \citep{qlora2023dettmers} with 8-bit quantization, LoRA $\alpha = 16$, and LoRA config $r = 64$. We trained for 3 epochs with a learning rate $lr = 1e^{-4}$, paged AdamW optimizer, cosine $lr$ scheduler, and an \textit{effective} batch size of 32 using gradient accumulation.

\section{Reasoning errors and their examples}
\label{sec:appendix_errors}

We randomly sampled and manually analyzed 80 model generated reasoning paths to identify the errors and understand the discrepancy in the GPT-4 and Llama-2 70B models. A collection of several errors, their examples in terms of reasoning steps, the datasets to which the generated paths belong and the explanation of the errors are provided in Table~\ref{tab:error_modes}.

\renewcommand{\minipagewidth}{\textwidth}
\begin{table*}[]
    \centering
    \small
    \begin{adjustbox}{max width=\textwidth}
    \begin{tabular}{l}
        \toprule
        \multicolumn{1}{c}{\textbf{Errors examples and explanations}} \\
        \midrule
        \begin{minipage}{\minipagewidth} \textbf{Error}: Reverse Step Merge \textbf{Dataset:} \texttt{SpaRP-S-PS2}(StepGame) \\
        \textbf{Reasoning Steps:} \textit{Step 7:} From step 4 and 6, we can infer that \textbf{I} is 2 unit above and 2 unit right of \textbf{W}. \textit{Step 9:} From step 8, we can infer that \textbf{Q} is right of \textbf{W}. \textit{Step 10:} From step 7 and 9, we can infer that \textbf{I} is 2 unit above and \textcolor{red}{3 unit right} of \textbf{Q}. \\
        \textbf{Explanation:} Requires composition from I to W to Q. Step 9 is a reverse step in this reasoning chain causing composition error.\end{minipage} \\
        \midrule
        \begin{minipage}{\minipagewidth} \textbf{Error:} Copied, not composed \textbf{Dataset:} \texttt{SpaRP-S-PS2} (StepGame) \\
        \textbf{Reasoning Steps:} \textit{Step 6:} From step 4 and 5, we can infer that \textbf{A} is below and right of \textbf{S}. \textit{Step 7:} From the context, \textbf{S} is \textcolor{red}{left} of \textbf{M}. \textit{Step 8:} From step 6 and 7, we can infer that \textbf{A} is below and right of \textbf{M}. \\
        \textbf{Explanation:} The relation from S to M is not used in composition, instead, the relation from A to S is copied for the merge step A to S to M.\end{minipage} \\
        \midrule
        \begin{minipage}{\minipagewidth} \textbf{Error:} Incorrect Composition \textbf{Dataset:} \texttt{SpaRP-S-PS2} (StepGame) \\
        \textbf{Reasoning Steps:} \textit{Step 11:} From step 8 and 10, we can infer that \textbf{Z} is above and 2 unit right of \textbf{L}. \textit{Step 13:} From step 12, we can infer that \textbf{L} is above and left of \textbf{J}. \textit{Step 14:} From step 11 and 13, we can infer that \textbf{Z} is \textcolor{red}{3 unit above} and right of \textbf{J}. \\
        \textbf{Explanation:} Quantitatively incorrect.\end{minipage} \\
        \midrule
        \begin{minipage}{\minipagewidth} \textbf{Error:} Merged without connection \textbf{Dataset:} \texttt{SpaRP-S-PS3} \\
        \textbf{Reasoning Steps:} \textit{Step 5:} From step 3 and 4, we can infer that \textbf{Y} is right of \textbf{L}. \textit{Step 14:} From step 12 and 13, we can infer that \textbf{C} is below and right of \textbf{K}. \textit{Step 15:} From step 5 and 14, we can infer that \textbf{C} is below and right of \textbf{L}. \\
        \textbf{Explanation:} No common entity between merged steps 5 and 14 which are 9 steps apart.\end{minipage} \\
        \midrule
        \begin{minipage}{\minipagewidth} \textbf{Error:} Non-composable Composition \textbf{Dataset:} \texttt{SpaRP-S-PS4} \\
        \textbf{Reasoning Steps:} \textit{Step 16:} From step 13 and 15, we can infer that \textbf{M} is below \textbf{X}. \textit{Step 18:} From step 17, we can say that \textbf{X} is above \textbf{A}. \textit{Step 19:} From step 16 and 18, we can infer that \textbf{M} is above \textbf{A}. \\
        \textbf{Explanation:} The reverse relations \{below, above\} are not composable under quantitatively unspecified (QU) criteria. \end{minipage} \\
        \midrule
        \begin{minipage}{\minipagewidth} \textbf{Error:} Reverse answer \textbf{Dataset:} \texttt{SpaRP-S-PS3} \\
        \textbf{Question:} What is the relation of the agent \textbf{W} to the agent \textbf{X}? \\
        \textbf{Answer Step:} \textit{Step 8:} From step 5 and 7, we can infer that \textbf{X} is above and left of \textbf{W}. \textit{Hence, the answer is} above, and left. \\
        \textbf{Explanation:} Directional relations are non-symmetric. Question is from W to X, while answer is from X to W.\end{minipage} \\
        \midrule
        \begin{minipage}{\minipagewidth} \textbf{Error:} Incorrect clock extraction \textbf{Dataset:}\texttt{SpaRP-S-PS3} \\
        \textbf{Context:} \textbf{P} is above \textbf{X} at 10 o'clock. \\
        \textbf{Extracted:} \textit{Step 6:} From the context, \textbf{X} is \textcolor{red}{below} \textbf{P}. \\
        \textbf{Explanation:} Correct extraction should be below and right.\end{minipage} \\
        \midrule
        \begin{minipage}{\minipagewidth} \textbf{Error:} Incomplete relation extraction \textbf{Dataset:} \texttt{SpaRP-S-PS4} \\
        \textbf{Context:} \textbf{I} is on the left side of and \textcolor{red}{below} \textbf{V}. \\
        \textbf{Extracted:} \textit{Step 2:} From the context, I is left of V. \\
        \textbf{Explanation:} Incomplete or partial extraction, below wasn't picked.\end{minipage} \\
        \midrule
        \begin{minipage}{\minipagewidth} \textbf{Error:} Incorrect relation extraction \textbf{Dataset:} \texttt{SpaRP-S-PS1} (\textsc{SpaRTUN}) \\
        \textbf{Context:} \textbf{Box EEE} \textit{\textcolor{ForestGreen}{has}} a \textbf{tiny white rectangle} and covers a midsize white diamond. \\
        \textbf{Extracted:} \textit{Step 7:} It is given that the \textbf{tiny white rectangle} is \textit{\textcolor{red}{inside and touching}} the \textbf{box EEE}. \\
        \textbf{Explanation:} \textit{Has} only means inside.\end{minipage} \\
        \midrule
        \begin{minipage}{\minipagewidth} \textbf{Error:} Incorrect reversal \textbf{Dataset:} \texttt{SpaRP-S-PS1} (\textsc{SpaRTUN}) \\
        \textbf{Reasoning Steps:} \textit{Step 1:} It is given that the \textbf{block AAA} \textit{\textcolor{ForestGreen}{contains}} the \textbf{medium yellow square}. \textit{Step 2:} From step 1, we can infer that the \textbf{medium yellow square} is \textit{\textcolor{red}{inside and touching}} the \textbf{block AAA}. \\
        \textbf{Explanation:} Reverse of \textit{contains} should only be inside.\end{minipage} \\
        \bottomrule
    \end{tabular}
    \end{adjustbox}
    \caption{Observed errors and their examples in the model generated reasoning paths. Only the relevant steps are shown.}
    \label{tab:error_modes}
\end{table*}